\documentclass[letterpaper]{article} % DO NOT CHANGE THIS
\usepackage{aaai2026}  % DO NOT CHANGE THIS, for camera ready
\usepackage{times}  % DO NOT CHANGE THIS
\usepackage{helvet}  % DO NOT CHANGE THIS
\usepackage{courier}  % DO NOT CHANGE THIS
\usepackage[hyphens]{url}  % DO NOT CHANGE THIS
\usepackage{graphicx} % DO NOT CHANGE THIS
\usepackage{natbib}  % DO NOT CHANGE THIS AND DO NOT ADD ANY OPTIONS TO IT
\usepackage{caption} % DO NOT CHANGE THIS AND DO NOT ADD ANY OPTIONS TO IT
\usepackage[utf8]{inputenc} % allow utf-8 input
\usepackage[T1]{fontenc}    % use 8-bit T1 fonts
\usepackage{booktabs}
\usepackage[accsupp]{axessibility}  % Improves PDF readability for those with disabilities.
\usepackage[longtable]{multirow}
\usepackage{makecell}
\usepackage{float}
\usepackage{icomma}
\usepackage{tabulary}
\usepackage{xcolor}
\usepackage{adjustbox}
\usepackage{caption}
\usepackage{subcaption} % for figure 1, subfigure
\usepackage{enumitem}
\definecolor{gray}{rgb}{0.3,0.3,0.3}
\usepackage{pifont}% http://ctan.org/pkg/pifont
\newcommand{\cmark}{\ding{51}}%
\newcommand{\xmark}{\ding{55}}%
\usepackage{tabularx}
\usepackage{adjustbox}

\usepackage{newfloat}
\usepackage{listings}
\DeclareCaptionStyle{ruled}{labelfont=normalfont,labelsep=colon,strut=off} % DO NOT CHANGE THIS
\floatstyle{ruled}
\newfloat{listing}{tb}{lst}{}
\floatname{listing}{Listing}
\newcommand{\ie}[1]{\textit{i.e.}{#1}}
\newcommand{\eg}[1]{\textit{e.g.}{#1}}

\usepackage{hyperref} 
\definecolor{cvprblue}{rgb}{0.21,0.49,0.74}

\hypersetup{
    colorlinks=true,
    linkcolor=red,
    filecolor=magenta,      
    urlcolor=magenta,
    citecolor=cvprblue
}

\title{
Hybri\textcolor{red}{D}LA: Hybrid Generation for Document Layout Analysis
}
\author{
Yufan Chen\textsuperscript{\rm 1},
Omar Moured\textsuperscript{\rm 1},
Ruiping Liu\textsuperscript{\rm 1},
Junwei Zheng\textsuperscript{\rm 1},
Kunyu Peng\textsuperscript{\rm 1},\\
Jiaming Zhang\textsuperscript{\rm 2,}\thanks{Corresponding author.},
Rainer Stiefelhagen\textsuperscript{\rm 1}
}
\affiliations{
    \textsuperscript{\rm 1} Institute for Anthropomatics and Robotics, Karlsruhe Institute of Technology\\
    \textsuperscript{\rm 2}School of Artificial Intelligence and Robotics, Hunan University\\
    \{yufan.chen, omar.moured, ruiping.liu, junwei.zheng, kunyu.peng, rainer.stiefelhagen\}@kit.edu,~jiamingzhang@hnu.edu.cn
}

\begin{document}

% \maketitle

% ---- For multiple figures 
\twocolumn[{%
\renewcommand\twocolumn[1][]{#1}%
\maketitle
\vspace{-3em} % added
\begin{center} 
    \captionsetup{type=figure}
    \setcounter{figure}{-1}
    \begin{subfigure}[t]{0.69\textwidth}
        \setcounter{figure}{1}
        \centering
        \includegraphics[width=\linewidth]{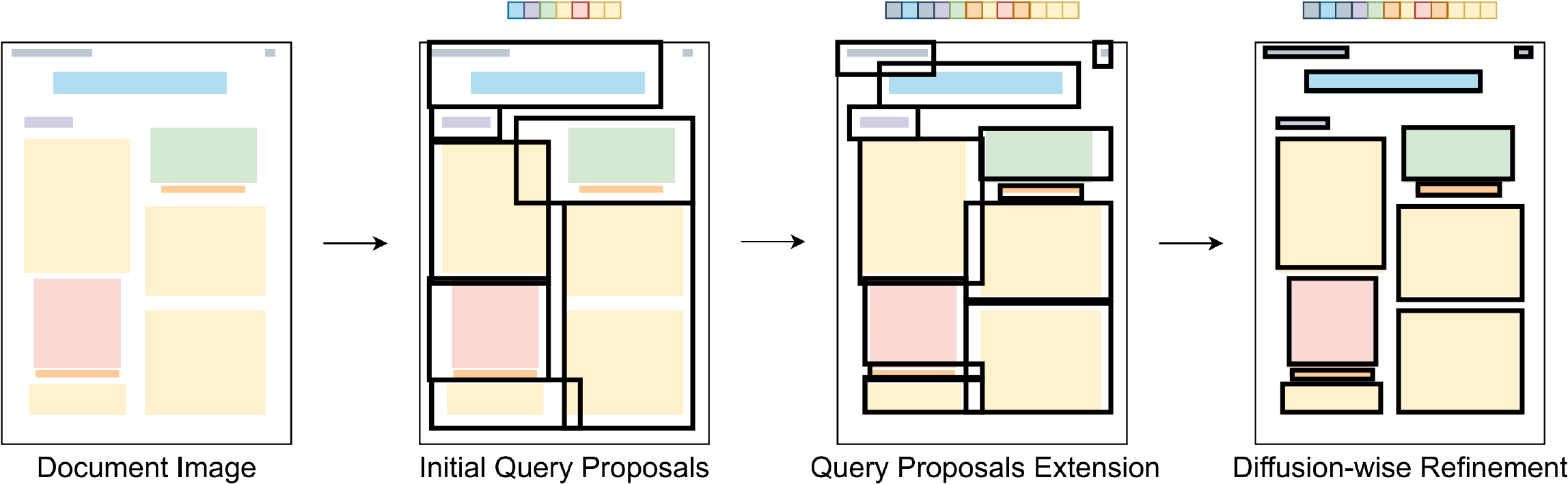}
        \caption{Representative paradigms for HybriDLA method}\label{fig1-a}
    \end{subfigure} \hspace{4pt}
    \begin{subfigure}[t]{0.24\textwidth}
        \centering
        \includegraphics[width=\linewidth]{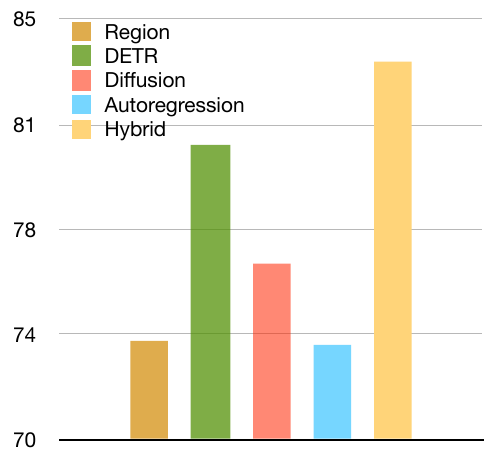}
        \caption{Performance on DocLayNet}\label{fig1-b}
    \end{subfigure}
    \setcounter{figure}{0}
    % \vspace{-1em} 
    \captionof{figure}{(a) The progressive prediction behaviors of representative paradigms in our proposed HybriDLA ensure successive intermediate outputs obtained during the forward pass from left to right. (b) On the DocLayNet~\cite{pfitzmann2022doclaynet} dataset, we compare the performance (mAP in \%) of the best models from five different DLA methods, \ie, traditional region-based method, DETR-based method, diffusion-based method, autoregressive method, and our proposed HybriDLA method.}
    \label{fig1:concept}
\end{center}

}]

{   % by-hand insert a footnote at page bottom, because the conflict between \twocolumn and footnote. 
 \renewcommand{\thefootnote}{\fnsymbol{footnote}}
 \footnotetext[1]{Corresponding author.}
 \footnotetext[0]{Copyright © 2026, Association for the Advancement of Artificial Intelligence (www.aaai.org). All rights reserved.}
}

\begin{abstract}
Conventional document layout analysis (DLA) traditionally depends on empirical priors or a fixed set of learnable queries executed in a single forward pass. While sufficient for early‐generation documents with a small, predetermined number of regions, this paradigm struggles with contemporary documents, which exhibit diverse element counts and increasingly complex layouts. To address challenges posed by modern documents, we present \textbf{HybriDLA}, a novel generative framework that unifies diffusion and autoregressive decoding within a single layer. The diffusion component iteratively refines bounding‐box hypotheses, whereas the autoregressive component injects semantic and contextual awareness, enabling precise region prediction even in highly varied layouts. To further enhance detection quality, we design a multi‐scale feature‐fusion encoder that captures both fine‐grained and high‐level visual cues. This architecture elevates performance to \textbf{83.5\%} mean Average Precision (mAP). Extensive experiments on the DocLayNet and M$^{6}$Doc benchmarks demonstrate that \textbf{HybriDLA} sets a state-of-the-art performance, outperforming previous approaches. All data and models will be made publicly available at \url{https://yufanchen96.github.io/projects/HybriDLA}. 
\end{abstract}	

\section{Introduction}
\label{sec:intro}
Understanding the layout of a document image is a fundamental task in document analysis. Accurate document layout analysis (DLA) is crucial for downstream applications, \eg, document understanding and information extraction. However, a persistent challenge in DLA is the extreme variability in the number and arrangement of layouts across diverse documents. A simple form might contain only a handful of text fields, while a dense scientific article page could contain dozens of paragraphs, figures, and tables. This variability makes it strenuous for approaches that assume a predefined number of object queries to perform well across all cases. 

Early layout analysis methods as cascade R-CNN~\cite{cai2018cascade}, often treat the problem similarly to generic object detection, using a limited set of proposals or queries to find layout elements on a single document page. Transformer-based detectors, \eg, DETR, adopt a predetermined number of learnable queries to predict all layout elements. While effective on scenes with relatively consistent object counts, fixed-query methods struggle when real layout elements, which could range from 2 to 200, greatly differ from the predefined number. If fewer queries are used, the model misses layouts that need to be detected; if a larger preset number is used, it must handle useless ``no object'' queries, leading to inefficiency and potential disturbance. This rigidity contrasts with how humans approach reading a document, \ie, we do not decide in advance exactly how many items we will find. Instead, we first scan the page to identify major regions, then gradually delve into each region to find sub-components, adjusting our expectations.

Motivated by this human reading strategy~\cite{furnas1986fisheye, liu2005readingbehavior}, we present the hybrid generation method for document layout analysis, \ie, \textbf{Hybri\textcolor{red}{D}LA}, a unified transformer-based model that, for the first time, explicitly simulates the human-like coarse-to-fine reading strategy within a single end-to-end trainable architecture. Our model dynamically adjusts the number of object queries hierarchically, allowing it to handle documents with widely varying numbers of elements efficiently. As illustrated in Figure~\ref{fig1:concept}, HybriDLA initially uses only a small set of object queries to obtain a rough sketch of the page layout. Each of these initial queries attends to a broad region of the page, potentially spanning multiple actual elements. Then, in subsequent decoding layers, the model progressively increases its resolution: it refines the localization of these coarse regions and spawns new queries to delve into detailed regions. During this continually iterative process, each decoder layer focuses on a finer level of granularity than the previous one. By the final layer, the model has transitioned from a coarse overview to a detailed, precise identification of every individual layout element on the page of the document image. 

To enable this hierarchical expansion and refinement, HybriDLA employs the hybrid generative decoding mechanism that combines the strengths of diffusion-based and autoregressive modeling. These two components work in tandem at each stage of the decoder. The diffusion-based part continuously refines the spatial coordinates of all current queries, analogous to fine-tuning the attention on a region for precise boundaries. Meanwhile, the autoregressive part handles the semantic and structural aspect, \ie, it determines if a query represents multiple elements that should be further split. If further detail is needed, the autoregressive decoder generates new queries for the next layer, targeting the sub-elements within a coarse region. 

To summarize, our contributions are as follows:
\begin{itemize}[leftmargin=0.8cm]
    \item We propose \textbf{HybriDLA}, a novel document layout analysis model that, for the first time, integrates a human-inspired coarse-to-fine querying strategy. 
    \item HybriDLA couples the feature fusion encoder with the hybrid generative decoder that unifies autoregressive query expansion and diffusion‑based refinement, enabling precise layout analysis.  
    \item Extensive experiments on DocLayNet and M$^6$Doc demonstrate the state‑of‑the‑art performance of our proposed HybriDLA method. 
\end{itemize}

\section{Related Work}
\label{related_work}

\subsection{Document Layout Analysis}
Document Layout Analysis (DLA) plays a crucial role in comprehending the structure and content organization within documents. Both traditional machine learning techniques~\cite{diem2011text,garz2010detecting} and modern deep learning approaches~\cite{long2022towards,gemelli2022doc2graph,peng2022ernie,zhu2022towards,coquenet2023dan,yang2022transformer} have witnessed substantial advancements with the availability of diverse datasets and benchmarks~\cite{li2020docbank,shihab2023badlad,zhong2019publaynet,shen2020large,moured2023line}. 
DLA has been explored using single-modal framework Faster R-CNN~\cite{ren2017FasterRcnn}, Mask R-CNN~\cite{He_2017_MaskRcnn}, and DocSegTr~\cite{biswas2022docsegtr}, as well as multi-modal models including LayoutLMv3~\cite{huang2022layoutlmv3} and DiT~\cite{li2022dit}. Furthermore, text grid-based techniques~\cite{zhang2021vsr} demonstrate the effective fusion of textual layout with visual cues. More recently, transformer-based architectures~\cite{coquenet2023dan, yang2022transformer, cheng2023m6doc, tang2023unifying, li2022dit, arroyo2021variational, wang2022lilt} have gained prominence in this field. Self-supervised pretraining methods~\cite{xu2020layoutlm, xu-etal-2021-layoutlmv2, li2021structext, Appalaraju_2021_DocFormer,Luo2022BiVLDocBV,huang2022layoutlmv3,luo2023geolayoutlm}, such as DocFormer~\cite{Appalaraju_2021_DocFormer} and LayoutLMv3~\cite{huang2022layoutlmv3}, have also attracted significant interest.  Zhang~\textit{et al.}~\cite{zhang2025understand} propose a unified feature-conductive end-to-end document image translation framework.
Chen~\textit{et al.}~\cite{chen2025graph} propose graph-based document layout analysis focusing on cross-page long-range document analysis. Zhang~\textit{et al.}~\cite{zhang2025sail} propose SAIL, a finetuning-free method for Document Information Extraction that leverages textual and layout similarity to guide LLMs, achieving strong performance without full training. Shen~\textit{et al.}~\cite{shen2025proctag} introduces ProcTag, a process-based tagging method for evaluating document instruction data, and DocLayPrompt, a layout-aware prompting strategy. Constum~\textit{et al.}~\cite{constum2025daniel} presents DANIEL, combining layout analysis, handwriting recognition, and named entity recognition in handwritten documents using a language-model-based decoder.

\subsection{Generic Detection Models}
\noindent \textbf{DETR-Based Detection Model.}
DETR~\cite{carion2020detr} reframed object detection as a bipartite set prediction task, \ie, a fixed collection of $N$ learnable queries is fed to a transformer decoder with Hungarian matching that aligns the $N$ predictions with the ground-truth objects and supervises them with a permutation-invariant loss.  While conceptually elegant, vanilla DETR converges slowly and struggles to localize small objects because each query must attend the full feature map at every decoder layer. Deformable DETR~\cite{zhu2021deformableDETR} tackles this by sampling a sparse set of keypoints around each reference point, reducing quadratic attention to linear cost and accelerating convergence. DINO~\cite{zhang2022dino}, which unifies the denoising, contrastive query selection, and look-forward twice refinement, achieves state-of-the-art accuracy with fewer epochs. RoDLA~\cite{chen2024rodla} suppresses local noise responses by inserting channel attention and average pooling into the self-attention mechanism, allowing it to remain stable under distortions. 

\noindent \textbf{Autoregressive Detection Model.}
Pix2Seq~\cite{chen2021pix2seq} pioneered a language-model view of detection: the model serializes each object as a sequence token $\langle{x}\,{y}\,{w}\,{h}\,{c}\rangle$ and generates tokens autoregressively with teacher forcing. Pix2Seq-v2~\cite{chen2022unified} extends this formulation to a unified multitask framework, improves box precision with relative coordinate encoding, and decodes up to 300 objects per image on COCO. While autoregressive decoding gracefully accommodates sequences of arbitrary length and captures rich inter-object dependencies, its computational cost grows linearly with the number of targets. 

\noindent \textbf{Diffusion‑based Detection Model.}
Inspired by score-based generative modeling, DiffusionDet~\cite{chen2022diffusiondet} samples $N$ Gaussian noise boxes at time step $T$ and trains the network to denoise them towards ground-truth boxes through $T$ steps.  During inference, only four to seven predictor steps are executed with a learned variational sampler, achieving strong detection performance and ensemble-like robustness.  Despite iterative quality gains, diffusion detectors still rely on a prefixed pool of initial proposals.

\section{Hybri\textcolor{red}{D}LA Framework}
\label{sec:method}
\subsection{Pipeline Overview} 
\begin{figure}
    \centering
    \includegraphics[width=0.99\linewidth]{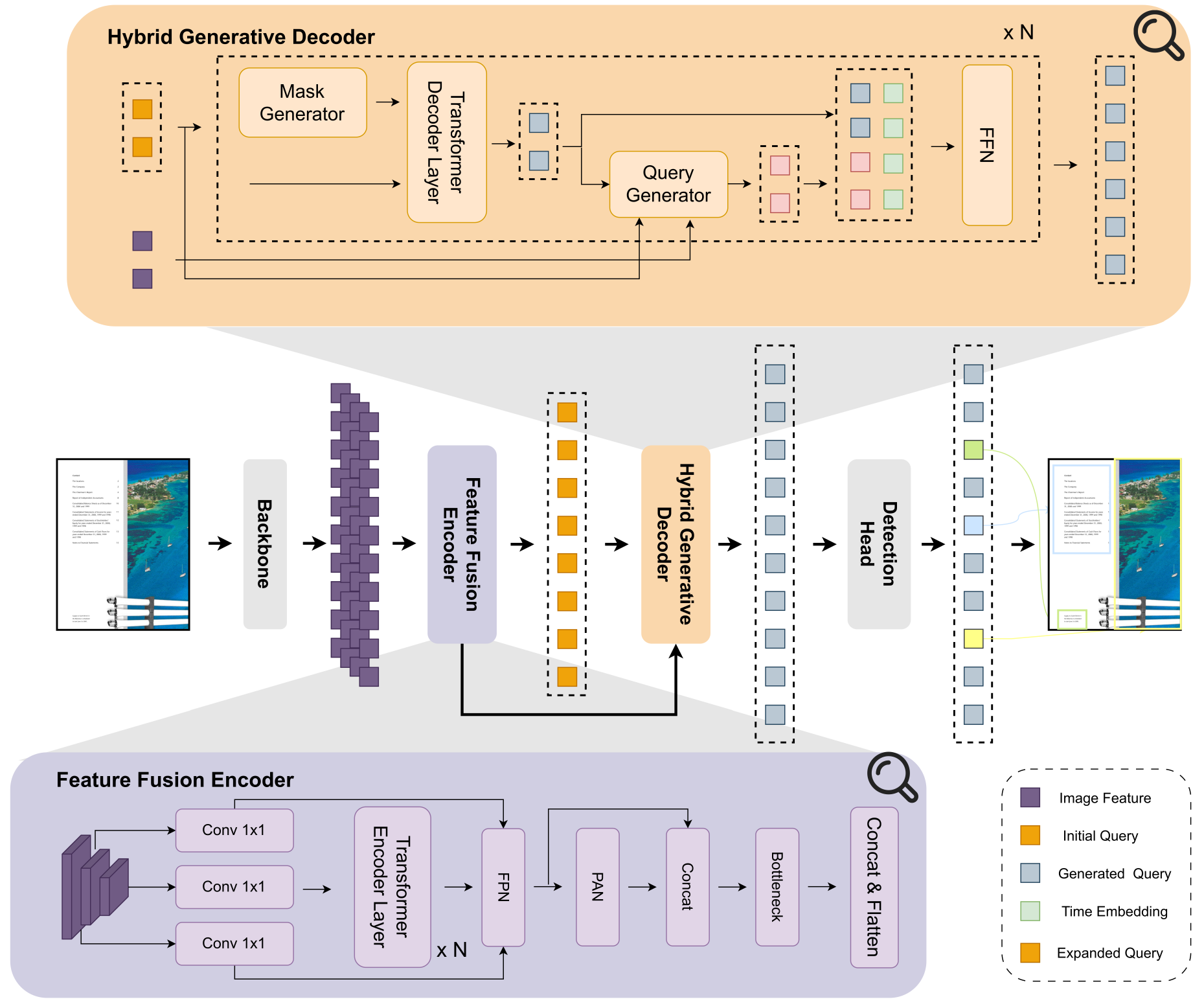}
    \caption{\textbf{Overview of the HybriDLA architecture.} The framework consists of a feature fusion encoder and a hybrid generative decoder. The encoder aggregates multi-scale visual features via convolutional and transformer layers, producing a layout-aware representation. The decoder operates in two mechanisms: it performs autoregressive query expansion to propose hierarchical layout regions, then applies a diffusion-style refinement with residual correction to denoise and adjust spatial predictions. Auxiliary queries and intermediate supervision facilitate convergence. This coarse-to-fine pipeline enables precise and adaptive generation of layout detection results.}
    \label{fig:HybridDLA}
\end{figure}
The HybriDLA framework introduces a human-inspired coarse-to-fine layout generation pipeline, integrating a top-down coarse prediction with a bottom-up refinement. As shown in Figure~\ref{fig:HybridDLA}, the HybriDLA framework follows a two-stage, hierarchical generation pipeline composed of a multi-scale feature fusion encoder and a hybrid generative decoder. In the first stage, the encoder processes the input document image features, which are extracted through the backbone, to produce coarse layout region predictions that capture the approximate locations and extents of major content blocks. These coarse predictions serve as high-level layout priors rather than final outputs. In the second stage, the hybrid decoder utilizes these coarse region priors, along with the rich feature representations of encoders, to refine them into a final set of layout elements with precise boundaries and semantic labels. This two-tier decoding strategy mirrors how a human would parse a page, first sketching high-level regions, then focusing on details. By combining top-down and bottom-up approaches, HybriDLA achieves high performance on accuracy in analysis. In the following, we will provide an in-depth description of these modules.

\subsection{Feature Fusion Encoder}
The feature fusion encoder is designed as a unified hierarchical representation extraction module that transforms a set of multi-scale feature maps ${F}_{l=1}^L$ into a single spatial-aware representation $G$. In essence, this encoder combines information from different scales in an organized, multi-stage manner so that the resulting representation $G$ captures the informative layout context and element content of documents.

\noindent \textbf{Local Feature Encoding}. For each input feature map $F_l$ (at scale $l$), the encoder first performs a local encoding to produce an enriched feature map $H_l = \phi(F_l)$. The local feature encoding function $\phi$ typically includes a combination of self-attention and convolutional operations applied within the module. The self-attention component enables the model to capture long-range dependencies and relationships within the feature map, complementing the convolution, which focuses on local patterns. In practice, this means $\phi$ can learn intra-scale patterns, \eg, global information such as textures and part-whole relationships that might span across the spatial extent of $F_l$. The outcome is that each $H_l$ is a refined version of $F_l$ at scale $l$. It preserves important fine-grained details of that scale while also embedding rich contextual information. This step ensures that, before any cross-scale fusion, features in each scale are locally enhanced.

\noindent \textbf{Cross-Scale Fusion}. After obtaining the set of enhanced features $H_{l=1}^L$, the encoder then applies an attention-based fusion function $\Psi$ to merge information across these scales. The cross-scale fusion $\Psi({H_l})$ is designed to perform adaptive information exchange between different feature scales. In practice, this involves a cross-scale attention where features at one focused scale integrate information from features at another scale. This allows the fine-detailed feature map to incorporate global context from a low-scale feature map, and vice versa, a coarse feature to be informed about fine details from a low-scale map. Additionally, lateral convolutional layers are used to align and combine feature maps from different scales. These lateral convolutions ensure that features from various scales are projected into a common space and integrated smoothly, allowing the network to learn a coherent, multi-scale representation.

Through local feature encoding and the cross-scale fusion module, the resulting representation $G = \Psi(H_{l=1}^L)$ is a hierarchical feature embedding which provides a global context understanding of the document layout while still retaining the necessary detail for precise prediction tasks. The representation $G$ could function as the initial coarse proposals of the important layout elements, which guide subsequent decoder processing. Besides, $G$ could also act as a source of multi-scale information for the hybrid generative decoder. Concretely, the decoder’s generated and expanded queries are guided by the learned feature information encoded in $G$.

\subsection{Hybrid Generative Decoder}
The decoder of HybriDLA is a hybrid generative module that combines an autoregressive query expansion mechanism with a diffusion-based refinement process. This design enables the adaptive generation of layout elements, transitioning from coarse proposals to fine-grained results within these two mutually complementary mechanisms.

\noindent \textbf{Autoregressive Query Expansion (AQE)}. We formulate the query expansion mechanism as an autoregressive generative process operating over spatial layout regions. Given an extracted document image feature denoted $X$, the model defines a probability distribution over a variable-length sequence of region queries $Q = (q_1, q_2, \dots, q_N)$. Each query $q_t$ represents a candidate region with associated spatial and semantic parameters, and is generated conditionally on the image feature and all previously generated queries. Formally, the joint distribution over the query sequence factorizes as:
\begin{equation}
\begin{aligned}
        P(Q \mid X) &= P(q_1, q_2, \ldots, q_N, \mathtt{EOS} \mid X) \; \\
                    &=\; \prod_{t=1}^{N} P(q_t \mid X,\; q_{1:t-1}) \;\cdot\; P(\mathtt{EOS} \mid X,\; q_{1:N}) 
\end{aligned}
\end{equation}
where $\mathtt{EOS}$ is a special end-of-sequence token that terminates the process. This formulation implies that at each step $t$, a learned proposal distribution $P(q_t \mid X, q_{<t})$, which is parameterized by the decoder layer, proposes the next query given the current context, and eventually emits $\mathtt{EOS}$ to adaptively halt expansion. The decoder layer can be viewed as a parametric function $D_{\theta,t}$ at iteration $t$ that operates over the growing query set. At each iteration step $t$, $D_{\theta,t}$ conditions on the input $X$ and the generated queries ${q_1,\dots,q_{t-1}}$ from the previous step, and further conditions the next conditional distribution for the new query $q_t$. In this way, the model autoregressively “queries the queries,” \ie, it conditions on its own past outputs to decide future ones. During this generative process, each $q_t$ is generated in contextual content. 
The cardinality $N$ of the query set is not fixed a priori but is controlled in a data-dependent manner by the learned stopping criterion $P(\mathtt{EOS} \mid X, q_{1:N})$. Sequentially, after every new set of queries has been proposed, the decoder layer assesses whether the current set of queries sufficiently explores the whole layout of the input document image $X$. If there remain unexplained regions or high residual uncertain regions, the model will continue to propose new queries. Conversely, when the layout has been fully accounted for, the $\mathtt{EOS}$ token is emitted, terminating the expansion.

\noindent \textbf{Diffusion-based Refinement (DR)}. We model the layout refinement process as an implicit denoising operation, similar to a diffusion model but without any forward noise injection. Each decoder layer treats its input prediction as a noisy estimation and produces a residual correction to progressively remove errors. Formally, $\hat{y}^{(t)}$ denote the layout prediction after $t$ refinement steps, %i.e., decoder layers in , 
the update rule could be formulated as $\hat{y}^{(t+1)} = \hat{y}^{(t)} + \Delta^{(t)},$ where $\Delta^{(t)}$ is the predicted residual at step $t$. This residual formulation ensures that the model focuses on correcting deviations from the previous prediction, effectively denoising $\hat{y}^{(t)}$ to approach the accurate layout. During the iteratively refinement by decoder layers, the self-attention mechanism allows decoder queries to share contextual information, while cross-attention integrates relevant visual features, enabling precise localized updates. The feed-forward network then adjusts each query embedding, applying the learned residual $\Delta^{(t)}$ to yield the refined prediction $\hat{y}^{(t+1)}$. Through these attention and feed-forward operations, successive decoder layers systematically reduce spatial errors and converge toward a high-fidelity layout estimation. To train this refinement mechanism, we incorporate denoising queries and intermediate supervision. During training, a subset of decoder queries is initialized with perturbed ground-truth layouts, challenging the network to reconstruct the correct layout from a degraded version. The model is supervised with auxiliary loss at each decoder layer, which forces intermediate predictions $\hat{y}^{(t)}$ to stay close to the ground truth, and with auxiliary heads for fast convergence. These auxiliary training strategy guides the decoder to robustly remove noise and refine predictions at every step, ensuring theoretical accuracy and stable convergence.

By combining autoregressive query expansion with diffusion-based refinement, HybriDLA can flexibly accommodate varying document complexities. If a document with many small elements, more queries will be generated in the expansion stage, and the diffusion refinement will meticulously adjust each one. If the layout is simple, the model generates fewer queries and still refines them for precision. This progressive refinement approach allows the model to iteratively approach the truth, focusing on coarse structure first, then detail, like a human would~\cite{liu2005readingbehavior, furnas1986fisheye}.

\section{Experiments}

\subsection{Experiment Setup} 
\noindent \textbf{Dataset}.We conduct experiments on the DocLayNet~\cite{pfitzmann2022doclaynet} and M6Doc~\cite{cheng2023m6doc} datasets. DocLayNet is a large human-annotated document layout dataset containing 80,863 page images with labeled bounding-box annotations for 11 distinct layout classes, \ie, Caption, Footnote, Formula, List-item, Page-footer, Page-header, Picture, Section-header, Table, Text, Title. M6Doc is a recent multi-format, multi-type, multi-layout dataset comprising 9,080 document pages with fine-grained bounding-box annotations for 74 categories, totaling 237,116 annotated instances. All annotations in both datasets are provided as axis-aligned bounding boxes for each layout element.

\noindent\textbf{Evaluation Protocol.} We adopt mean Average Precision (mAP) as the primary evaluation metric, following the standard definition. Specifically, mAP is computed by averaging the Average Precision over a range of Intersection over Union (IoU) thresholds from 0.50 to 0.95 in 0.05 increments. This mAP summarizes detection accuracy across multiple overlap criteria, and we report it as an overall indicator of performance, \ie, higher mAP indicates better layout detection. All results are reported on the respective official setting of each dataset.

\noindent \textbf{Implementation Details.} We benchmark HybriDLA against a comprehensive list of analyzers reproduced under the same training setting as in Table~\ref{tab:sota_doclaynet}, which include 
\begin{itemize}
    \item \textbf{Traditional region‑based method}: Classical two‑stage pipelines remain the standard for layout analysis. We reproduce the strongest public variants with both convolutional and transformer backbones. These methods establish baselines and are still widely used in industry.
    \item \textbf{DETR-based method}: We cover the evolution from vanilla DETR to domain‑specialized DLAFormer with both convolutional and Transformer backbones. Set-prediction makes DETR variants the most competitive contemporary baselines.
    \item \textbf{Diffusion-based method}: DiffusionDet~\cite{chen2022diffusiondet} represents the emerging class of generative detectors. We evaluate on both convolutional and transformer backbones to examine whether coarse‑to‑fine refinement benefits complex page structures.
\item \textbf{Autoregressive method}: We include Pix2Seq~\cite{chen2021pix2seq} with transformer backbone. These methods sequentially predict objects as token strings, offering an alternative that unifies detection and language modeling.

    \item \textbf{Hybrid method (ours)}: We test HybriDLA method with 5 backbones to systematically validate the generality. 
\end{itemize}

All backbones are initialized from the official weights, while analyzers are trained from scratch. Every model is optimized with a batch size of 40, except for DLAFormer, retrained in the same setting to ensure comparability.

\subsection{Experiment Results on DocLayNet}
\begin{table}[t]
\centering
\caption{Experiment results on DocLayNet dataset.  
V, L, and T denote the Visual, Layout, and Textual modalities.}
\label{tab:sota_doclaynet}
\setlength{\tabcolsep}{3pt}      %
\renewcommand{\arraystretch}{.99}% 
\begin{adjustbox}{width=\linewidth}
\begin{tabularx}{1.1\linewidth}{llc*{3}{>{\centering\arraybackslash}X}c}
\toprule[1.2pt]
\multirow{2}{*}{\textbf{Backbone}}
&\multirow{2}{*}{\textbf{Detector}}
&\multirow{2}{*}{\textbf{\#Params}}
&\multicolumn{3}{c}{\textbf{Modality}}
&\multirow{2}{*}{\textbf{mAP}}\\
&&& V & L & T &\\
\midrule \midrule
\multicolumn{7}{l}{\textbf{\emph{Region-based Method}}}\\
ResNet-101 & Mask R-CNN      & 63M &\cmark&\xmark&\xmark& 73.5\\
ResNet-101 & Faster R-CNN    & 61M &\cmark&\xmark&\xmark& 73.4\\
ResNet-101 & DocSegTr        & 103M &\cmark&\xmark&\xmark& 69.3\\
DiT        & Cascade R-CNN   & 304M &\cmark&\xmark&\xmark& 62.1\\
\textcolor{gray!30}{LayoutLMv3} & \color{gray!30}{Cascade R-CNN} & \textcolor{gray!30}{368M} &\textcolor{gray!30}{\cmark}&\textcolor{gray!30}{\cmark}&\textcolor{gray!30}{\cmark}& \textcolor{gray!30}{75.1}\\
\midrule
\multicolumn{7}{l}{\textbf{\emph{DETR-based Method}}}\\
InternImage & DETR              & 352M &\cmark&\xmark&\xmark& 74.3\\
InternImage & Deformable DETR   & 353M &\cmark&\xmark&\xmark& 76.1\\
InternImage & DINO              & 358M &\cmark&\xmark&\xmark& 75.7\\
InternImage & Co-DINO           & 363M &\cmark&\xmark&\xmark& 77.2\\
Swin-L      & SwinDocSegmenter  & 223M &\cmark&\xmark&\xmark& 76.9\\
InternImage & RoDLA             & 323M &\cmark&\xmark&\xmark& 80.5\\
\textcolor{gray!30}{ResNet-50} & \textcolor{gray!30}{DLAFormer} & \textcolor{gray!30}{--} &\textcolor{gray!30}{\cmark}&\textcolor{gray!30}{\cmark}&\textcolor{gray!30}{\cmark}& \textcolor{gray!30}{83.8}\\
\midrule
\multicolumn{7}{l}{\textbf{\emph{Diffusion-based Method}}}\\
ResNet-50 & DiffusionDet  & 111M &\cmark&\xmark&\xmark& 73.7\\
Swin-L    & DiffusionDet  & 283M &\cmark&\xmark&\xmark& 76.3\\
\midrule
\multicolumn{7}{l}{\textbf{\emph{Autoregressive Method}}}\\
ViT-L & Pix2Seq      & 341M &\cmark&\xmark&\xmark& 72.5\\
FIT-L & Pix2Seq  & 370M &\cmark&\xmark&\xmark& 73.4\\
\midrule
\multicolumn{7}{l}{\textbf{\emph{Hybrid Method}}}\\
ResNet-50    & Ours & 95M &\cmark&\xmark&\xmark& 74.4\\
ResNet-101   & Ours & 114M &\cmark&\xmark&\xmark& 76.9\\
ViT-L        & Ours & 385M &\cmark&\xmark&\xmark& 78.8\\
Swin-L       & Ours & 270M &\cmark&\xmark&\xmark& 80.4\\
InternImage  & Ours & 392M &\cmark&\xmark&\xmark& \textbf{83.5}\\
\bottomrule[1.2pt]
\end{tabularx}
\end{adjustbox}
\end{table}

Table~\ref{tab:sota_doclaynet} summarizes the detection performance of various methods on the DocLayNet~\cite{pfitzmann2022doclaynet} benchmark. Traditional region-based methods with only vision achieve less than 73.5\% mAP. ResNet-101~\cite{kaiming2015resnet} with Mask R-CNN~\cite{He_2017_MaskRcnn} reaches 73.5\% mAP, followed by Faster R-CNN~\cite{ren2017FasterRcnn} at 73.4\%. While self-attention integrated DocSegTr~\cite{biswas2022docsegtr} yields 69.3\%, a specialized document-transformer backbone DiT~\cite{li2022dit} with Cascade R-CNN~\cite{cai2018cascade} attains only 62.1\%. Incorporating textual input provides a modest boost to LayoutLMv3~\cite{huang2022layoutlmv3}, improving to 75.1\% mAP, which indicates the benefit of multimodal cues over purely visual baselines. 

In contrast, DETR-based methods achieve higher accuracy on DocLayNet~\cite{pfitzmann2022doclaynet}, with mAP more than 73\%. Vanilla DETR~\cite{carion2020detr} with InternImage~\cite{wang2023internimage} backbone obtains 74.3\% mAP, while Deformable DETR~\cite{zhu2021deformableDETR} and DINO~\cite{zhang2022dino} reach 76.1\% and 75.7\% mAP, respectively. Co-DINO~\cite{zong2022codetr} further improves to 77.2\% mAP, and a Swin-based SwinDocSegmenter~\cite{banerjee2023swindocsegmenter} achieves 76.9\% mAP. The strongest vision-only DETR-based method is RoDLA~\cite{chen2024rodla} at 80.5\% mAP, representing a substantial improvement. Besides, DiffusionDet~\cite{chen2022diffusiondet} yields up to 76.3\% mAP with Swin Transformer~\cite {liu2021swin} backbone, while autoregressive model Pix2Seq~\cite{chen2021pix2seq} with ViT~\cite{dosovitskiy2020vit} backbone reach 72.5\% mAP.

Compared to these baseline models, our HybriDLA model obtains a state-of-the-art performance among vision-only document layout analysis models. As shown in the bottom section of Table~\ref{tab:sota_doclaynet}, the hybrid approach outperforms prior methods across diverse backbones. With the InternImage~\cite{wang2023internimage} backbone, HybriDLA achieves 83.5\% mAP, which is the best result for vision-only layout analysis model. This nearly matches the performance of 83.8\% mAP by DLAFormer~\cite{wang2024dlaformerendtoendtransformerdocument}, indicating that HybriDLA narrows the performance gap to multi-modal systems using only visual features. HybriDLA also yields an average ~3\% mAP gains with other backbones. Even with a smaller ResNet-50, HybriDLA achieves 74.4\%, essentially matching the DiffusionDet baseline. These consistent gains across different backbone models demonstrate the effectiveness and generality of our HybriDLA model.

\subsection{Experiment Results on M$^6$Doc}
\begin{table}[t]
\centering
\caption{Experiment results on M$^{6}$Doc dataset.}
\label{tab:sota_m6doc}
\setlength{\tabcolsep}{3pt}
\renewcommand{\arraystretch}{1.1}
\begin{adjustbox}{width=\linewidth}
\begin{tabularx}{1.1\linewidth}{llc*{3}{>{\centering\arraybackslash}X}c}
\toprule[1.2pt]
\multirow{2}{*}{\textbf{Backbone}} & 
\multirow{2}{*}{\textbf{Detector}} & 
\multirow{2}{*}{\textbf{\#Params}} &
\multicolumn{3}{c}{\textbf{Modality}} & 
\multirow{2}{*}{\textbf{mAP}} \\[2pt]
& & & V & L & T & \\
\midrule\midrule
\multicolumn{6}{l}{\textbf{\emph{Region-based Method}}}\\
ResNet-101 & Mask R-CNN        & 63M & \cmark & \xmark & \xmark & 61.9\\
ResNet-101 & Faster R-CNN      & 61M & \cmark & \xmark & \xmark & 62.0\\
ResNet-101 & DocSegTr          & 103M & \cmark & \xmark & \xmark & 60.3\\
DiT        & Cascade R-CNN     & 304M & \cmark & \xmark & \xmark & 70.2\\
\textcolor{gray!30}{LayoutLMv3} & \color{gray!30}{Cascade R-CNN} & \textcolor{gray!30}{368M} &\textcolor{gray!30}{\cmark}&\textcolor{gray!30}{\cmark}&\textcolor{gray!30}{\cmark}& \textcolor{gray!30}{64.3}\\
\midrule
\multicolumn{6}{l}{\textbf{\emph{DETR-based Method}}}\\
InternImage & DETR              &352M& \cmark & \xmark & \xmark & 54.2\\
InternImage & Deformable DETR   &353M& \cmark & \xmark & \xmark & 61.2\\
InternImage & DINO              &358M& \cmark & \xmark & \xmark & 66.8\\
Swin-L      & SwinDocSegmenter  &223M& \cmark & \xmark & \xmark & 47.1\\
InternImage & RoDLA             &323M& \cmark & \xmark & \xmark & 70.0\\
\midrule
\multicolumn{6}{l}{\textbf{\emph{Diffusion-based Method}}}\\
Swin-L & DiffusionDet          &283M& \cmark & \xmark & \xmark & 62.7\\
\midrule
\multicolumn{6}{l}{\textbf{\emph{Autoregressive Method}}}\\
ViT-L & Pix2Seq               &341M& \cmark & \xmark & \xmark & 54.9\\
FIT-L & Pix2Seg           &370M& \cmark & \xmark & \xmark & 54.9\\
\midrule
\multicolumn{6}{l}{\textbf{\emph{Hybrid Method}}}\\
ResNet-50   & Ours &95M& \cmark & \xmark & \xmark & 62.1\\
ResNet-101  & Ours &114M& \cmark & \xmark & \xmark & 64.7\\
ViT-L       & Ours &385M& \cmark & \xmark & \xmark & 68.6 \\
Swin-L      & Ours &270M& \cmark & \xmark & \xmark & 68.1 \\
InternImage & Ours &392M& \cmark & \xmark & \xmark & \textbf{71.4} \\
\bottomrule[1.2pt]
\end{tabularx}
\end{adjustbox}
\end{table}

Table~\ref{tab:sota_m6doc} presents the detection accuracy of various approaches on the challenging M$^{6}$Doc~\cite{cheng2023m6doc} dataset. For traditional region-based methods, ResNet-101~\cite{kaiming2015resnet} with Mask R-CNN~\cite{He_2017_MaskRcnn} achieves 61.9\% mAP, and Faster R-CNN~\cite{ren2017FasterRcnn} yields a similar 62.0\%. The self-attention-based DocSegTr~\cite{biswas2022docsegtr} attains 60.3\% mAP, slightly lower than the R-CNN models. Notably, the document-specific transformer backbone DiT~\cite{li2022dit} combined with Cascade R-CNN~\cite{cai2018cascade} stands out with a significantly higher 70.2\% mAP, indicating the benefit of pretraining on document layouts even without text input.  We also observe that incorporating textual modality provides only a modest boost on M$^{6}$Doc~\cite{cheng2023m6doc} dataset, the multimodal LayoutLMv3~\cite{huang2022layoutlmv3} with multimodal features only reaches 64.3\% mAP. 

Transformer-based DETR-style methods exhibit a wide range of results. The vanilla DETR~\cite{carion2020detr} with an InternImage~\cite{wang2023internimage} backbone reaching only 54.2\% mAP, likely due to the difficulty with a large number of classes and diverse layouts. Introducing multi-scale deformable attention significantly improves performance. Deformable DETR~\cite{zhu2021deformableDETR} rises to 61.2\% mAP, and DINO~\cite{zhang2022dino} achieves 66.8\% mAP. Interestingly, the SwinDocSegmenter~\cite{banerjee2023swindocsegmenter} underperforms on this dataset, with only 47.1\% mAP. Among DETR-based approaches, the recent RoDLA~\cite{chen2024rodla} obtains the highest result at 70.0\% mAP, nearly matching the DiT-based region method. 

Beyond DETR-style models, DiffusionDet~\cite{chen2022diffusiondet} achieves 62.7\% mAP, which is better than the initial DETR and R-CNN baselines but still below the SOTA-performing transformer methods. Autoregressive sequence prediction approaches perform less competitively, both Pix2Seq~\cite{chen2021pix2seq} and its improved variant with FIT~\cite{chen2023fit} achieve 54.9\% mAP. 

In contrast to the above baselines, our proposed HybriDLA approach establishes a state-of-the-art performance for vision-only document layout analysis on M$^{6}$Doc~\cite{cheng2023m6doc} dataset. As shown in the bottom section of Table~\ref{tab:sota_m6doc}, HybriDLA consistently outperforms prior methods across all backbone architectures. With the powerful InternImage backbone, HybriDLA achieves 71.4\% mAP and even outperforms the multimodal LayoutLMv3~\cite{huang2022layoutlmv3} by over 7\% mAP. With ResNet-101~\cite{kaiming2015resnet} backbone, HybriDLA reaches 64.7\%, which is a 2\% obvious improvement over traditional R-CNN methods on the same backbone. Similarly, with the ViT~\cite{dosovitskiy2020vit} and Swin Transformer~\cite{liu2021swin} backbones, HybriDLA yields 68.6\% mAP and 68.1\% mAP, respectively, considerably higher than the corresponding DETR-based and DiffusionDet~\cite{chen2022diffusiondet} results. These consistent gains across diverse backbone types demonstrate the effectiveness and generality of our hybrid approach. 

\subsection{Ablation Study}
To comprehensively evaluate our proposed HybriDLA method, we conducted ablation studies from five perspectives on the DocLayNet~\cite{pfitzmann2022doclaynet} dataset.

\begin{table}[t]
\centering
\caption{Ablation study of autoregressive query expansion (AQE) mechanism. Deformable stands for deformable attention mechanism, and DINO stands for improved denoising anchor boxes mechanism.}
\label{tab:ablation_ar}
\setlength{\tabcolsep}{4pt}      % column separation
\renewcommand{\arraystretch}{.99}% row height
\begin{adjustbox}{width=\linewidth}
\begin{tabularx}{\linewidth}{cccccc}
\toprule[1.2pt]
\textbf{Backbone} & \textbf{DETR} & \textbf{Deformable} & \textbf{DINO} & \textbf{AQE} & \textbf{mAP}\\
\midrule \midrule
ResNet-50 & \cmark & \xmark & \xmark & \xmark & 74.2 \\
ResNet-50 & \cmark & \xmark & \xmark & \cmark & 74.4 \\
ResNet-50 & \cmark & \cmark & \xmark & \xmark & 75.1 \\
ResNet-50 & \cmark & \cmark & \xmark & \cmark & 76.3 \\
ResNet-50 & \cmark & \cmark & \cmark & \xmark & 75.3 \\
ResNet-50 & \cmark & \cmark & \cmark & \cmark & 76.8 \\
\midrule
Swin-L & \cmark & \cmark & \cmark & \cmark & 77.2 \\
\bottomrule[1.2pt]
\end{tabularx}
\end{adjustbox}
\end{table}
\noindent \textbf{The effect of autoregressive query expansion (AQE) mechanism.}
As shown in Table~\ref{tab:ablation_ar}, enabling the AQE mechanism consistently improves detection performance across different model variants. 
Compared to a marginal mAP increase of the ResNet-50 DETR baseline, stronger DINO
DETR~\cite{zhang2022dino} raises mAP from 75.3\% to 76.8\%, indicating a marginal effect. This trend suggests that the benefits of AQE become more pronounced as the layout analyzer is more advanced, which shows the complementary nature of the proposed AQE mechanism. 

\begin{table}[t]
\centering
\caption{Ablation study on feature fusion encoder, diffusion-based refinement mechanism and backbone selection. FFE stands for feature fusion encoder, DE stands for normal deformable attention encoder, DR stands for diffusion-based refinement mechanism, and AQE stands for autoregressive query expansion mechanism.}
\label{tab:ablation_encoder}
\setlength{\tabcolsep}{3pt}      % column separation
\renewcommand{\arraystretch}{.99}% row height
\begin{adjustbox}{width=\linewidth}
\begin{tabularx}{\linewidth}{l c *{3}{>{\centering\arraybackslash}X} c}
\toprule[1.2pt]
\textbf{Backbone} & \textbf{Encoder} & \textbf{DR} & \textbf{AQE} & \textbf{mAP}\\
\midrule \midrule
%ResNet-50   & DE & \xmark & \cmark & 76.8 \\
ResNet-50   & DE & \cmark & \cmark & 76.2 \\
ResNet-50   & FFE & \xmark & \cmark & 74.4 \\
ResNet-50   & FFE & \cmark & \cmark & 74.4 \\
\midrule
%Swin-L      & DE & \xmark & \cmark & 77.2 \\
Swin-L      & DE & \cmark & \cmark & 78.1 \\
Swin-L      & FFE & \xmark & \cmark & 79.1 \\
Swin-L      & FFE & \cmark & \cmark & 80.4 \\
\midrule
ResNet-101  & DE & \cmark & \cmark & 76.6 \\
ResNet-101  & FFE & \xmark & \cmark & 76.1 \\
ResNet-101  & FFE & \cmark & \cmark & 76.9 \\
\midrule
ViT-L       & DE & \cmark & \cmark & 79.0 \\
ViT-L       & FFE & \xmark & \cmark & 77.2 \\
ViT-L       & FFE & \cmark & \cmark & 78.8 \\
\midrule
InternImage & DE & \cmark & \cmark & 82.4 \\
InternImage & FFE & \xmark & \cmark & 81.3 \\
InternImage & FFE & \cmark & \cmark & 83.5 \\
\bottomrule[1.2pt]
\end{tabularx}
\end{adjustbox}
\end{table}

\noindent \textbf{The effect of feature fusion encoder (FFE).}
Table \ref{tab:ablation_encoder} contrasts the proposed FFE with the standard deformable encoder (DE). On high-capacity, multi-scale backbones, the FFE delivers clear benefits. With Swin‑L, mAP rises from 77.2\% to 79.1\% even before any diffusion refinement mechanism, and reaches 80.4\% mAP once FFE is paired with DR, which contains 2.3\% mAP gains compared to DE with DR. FFE with InternImage~\cite{wang2023internimage} shows a similar trend, climbing from 82.4\% mAP to 83.5\% mAP. ResNet‑101~\cite{kaiming2015resnet} also benefits slightly. In contrast, smaller models see limited impact, \eg, ResNet‑50~\cite{kaiming2015resnet} drops from 76.2\% mAP to 74.4\% mAP without DR. These results indicate that the FFE is advantageous when the backbone provides context-rich features.

\noindent \textbf{The effect of diffusion-based refinement (DR) mechanism.} As shown in Table~\ref{tab:ablation_encoder}, integrating the DR mechanism yields a consistent performance uplift across all backbones with the sole exception of ResNet‑50~\cite{kaiming2015resnet}, whose performance remains unchanged irrespective of whether DR is employed. The ResNet-101~\cite{kaiming2015resnet} gains 0.8\% mAP improvement, and similarly, the mAP result of Swin Transformer~\cite{liu2021swin} increases from 79.1\% to 80.4\%. The magnitude of improvement varies with model capacity. Smaller backbones tend to gain relatively more, while larger models slightly improve, implying that a larger backbone may require more steps to eliminate residual noise while more detailed features are extracted from images. 

\noindent \textbf{The generalization of backbone selection.} 
Table~\ref{tab:ablation_encoder} also demonstrates monotonic scaling with model capacity, as the backbone network capacity increases, performance increases continuously without exception. The mAP results rise from 74.4\% with ResNet-50~\cite{kaiming2015resnet}, to 83.5\% with InternImage~\cite{wang2023internimage}, showing a clear upward trend. This monotonic improvement indicates that our approach leverages the representational strength of larger backbones, achieving higher accuracy at each step up in model size. Moreover, the HybriDLA framework exhibits remarkable stability across diverse types of backbones, maintaining stable results regardless of backbone types. 

\begin{table}[t]
\centering
\caption{Ablation study on the impact of the number of initial proposal queries and AQE queries.}
\label{tab:ablation_query_number}
\setlength{\tabcolsep}{3pt}      % column separation
\renewcommand{\arraystretch}{.99}% row height
\begin{adjustbox}{width=\linewidth}
\begin{tabularx}{\linewidth}{l >{\centering\arraybackslash}X >{\centering\arraybackslash}X >{\centering\arraybackslash}X}
\toprule[1.2pt]
\textbf{Backbone} & \textbf{\#Init.\ Q.} & \textbf{\#AQE Q.} & \textbf{mAP}\\
\midrule \midrule
ResNet-50     & 300 & 300 & 74.4 \\
ResNet-101    & 300 & 300 & 76.9 \\
ViT-L         & 300 & 300 & 78.8 \\
Swin-L        & 300 & 300 & 80.4 \\
InternImage   & 300 & 300 & 83.5 \\
\midrule
ResNet-50     & 300 &  30 & 77.4 \\
ResNet-101    & 300 &  30 & 76.9 \\
ViT-L         & 300 &  30 & 78.2 \\
Swin-L        & 300 &  30 & 78.2 \\
InternImage   & 300 &  30 & 80.7 \\
\midrule
ResNet-50     & 900 &  30 & 75.5 \\
ResNet-101    & 900 &  30 & 76.6 \\
ViT-L         & 900 &  30 & 79.0 \\
Swin-L        & 900 &  30 & 75.7 \\
InternImage   & 900 &  30 & 81.4 \\
\bottomrule[1.2pt]
\end{tabularx}
\end{adjustbox}
\end{table}

\noindent \textbf{The analysis for hyper-parameters.} As shown in Table~\ref{tab:ablation_query_number}, the number of initial queries and expanded autoregressive queries has a significant impact on detection performance. For smaller backbones, reducing the expanded queries from 300 to 30 does not degrade performance. In contrast, larger backbones benefit from more queries, cutting the expanded queries to 30 leads to notable drops in mAP. Moreover, increasing the initial queries from 300 to 900 produces inconsistent results. The highest-capacity models show slight improvements, while others suffer a decline. These results suggest that each model has an optimal query budget. Smaller models saturate with fewer queries, while larger models can leverage more queries for better performance.

\section{Conclusion}
In this work, we introduced HybriDLA, a hybrid approach to document layout analysis that addresses complex and diverse document layouts. HybriDLA employs a coarse-to-fine generation strategy via diffusion-based layout refinement and autoregressive query expansion mechanisms, effectively capturing both global layout context and fine-grained details. Empirical evaluations on datasets demonstrate that HybriDLA achieves state-of-the-art performance in document layout analysis. Moreover, the architecture-agnostic design of HybriDLA enables seamless integration with various backbone structures, illustrating the generality and broad applicability. However, the current framework relies solely on visual inputs, which constitutes a notable limitation of analysis accuracy. As future work, we intend to address this issue by incorporating multimodal features to guide the layout generation process for better performance.

\clearpage
\section*{Acknowledgments}

This work was supported in part by Helmholtz Association of German Research Centers, in part by the Ministry of Science, Research and the Arts of Baden-Württemberg (MWK) through the Cooperative Graduate School Accessibility through AI-based Assistive Technology (KATE) under Grant BW6-03, and in part by National Natural Science Foundation of China under Grant No. 62503166. This work was partially performed on the HoreKa supercomputer funded by the MWK and by the Federal Ministry of Education and Research, partially on the HAICORE@KIT partition supported by the Helmholtz Association Initiative and Networking Fund, and partially on bwForCluster Helix supported by the state of Baden-Württemberg through bwHPC and the German Research Foundation (DFG) through grant INST 35/1597-1 FUGG. 
 % added 
\bibliography{main}
% \clearpage
% \input{ReproducibilityChecklist}
\clearpage
\appendix

\section{Social Impact and Limitations}

\noindent \textbf{Social Impact.}
We demonstrate that the advancement of HybriDLA in document layout analysis can directly benefit several real-world domains. In digital archiving, our method helps preserve and digitize complex historical documents by accurately recognizing diverse layouts, which aids libraries and museums in making cultural heritage accessible online. It also enhances screen-reader accessibility by converting visually rich layouts into structured content flows, empowering visually impaired users to navigate documents that were previously inaccessible due to complex formatting. Similarly, in education, HybriDLA can automatically organize content into logical sections, figures, and captions, facilitating interactive and adaptive learning materials. Additionally, by accurately extracting and grouping fields in forms and administrative documents, our system can streamline data entry and information retrieval in business and e-governance applications. These use cases highlight HybriDLA’s potential for broad positive social impact through improved access to information and automation of document workflows.

\noindent \textbf{Limitations}
Despite its strengths, our approach has several limitations. First, HybriDLA relies on document images without leveraging textual context or OCR results, which means it might miss semantic cues that could disambiguate layout elements, \eg, distinguishing figure captions from body text purely by appearance. Second, the hybrid generation mechanism of our model comes with a high inference cost, which combines multiple analysis stages and generative components, making it computationally intensive and less practical for real-time or large-scale processing. This could limit deployment on edge devices or slow down analysis for massive archives. We are exploring several future directions to address these issues. For instance, incorporating multimodal inputs or metadata could enrich the model’s understanding beyond pure visuals, and optimizing the architecture or using model distillation can reduce inference overhead. These steps should make HybriDLA more practical and reliable for widespread adoption.

\section{Experiment Details}

\subsection{Evaluation Metric Details.}
We report \textbf{mean Average Precision (mAP)} following the COCO object‑detection evaluation protocol~\cite{lin2014coco}. For each class $c \in \mathcal{C}$ and IoU threshold $t \in \mathcal{T}$, we compute the Average Precision (AP) as the area under the precision–recall curve:

\begin{equation}
    \mathrm{AP}_{c,t}= \int_{0}^{1} \mathrm{Prec}_{c,t}(r)\,dr,
\end{equation}

where $\mathrm{Prec}_{c,t}(r)$ denotes the maximum precision obtained for recall $r$ after 101‑point interpolation used in the COCO API. The mAP we report averages AP over all classes and over the 10 IoU thresholds

\begin{equation}
    \mathcal{T}= \{0.50,0.55,\ldots,0.95\},
\end{equation}

yielding

\begin{equation}
    \mathrm{mAP}= \frac{1}{|\mathcal{C}|\,|\mathcal{T}|}\sum_{c\in\mathcal{C}}\sum_{t\in \mathcal{T}}\mathrm{AP}_{c,t}.
\end{equation}

This metric rewards detectors that are simultaneously accurate in classification and precise in localization across a spectrum of overlap criteria, and it is the standard for fair comparison in modern document layout analysis benchmarks.  All results in the main paper and supplementary material are computed with the official COCO~\cite{lin2014coco} evaluation script to ensure reproducibility.

\subsection{Implementation Details.}
To comprehensively benchmark both classic and state‑of‑the‑art object‑detection paradigms for document layout analysis (DLA), we trained 15 distinct backbone–detector combinations, which span region‑based, DETR‑based, diffusion‑based, and autoregressive methods on the DocLayNet and M6Doc datasets. In addition, we evaluated five backbone variants of our proposed HybriDLA detector, enabling a direct comparison against existing SOTA methods. The architectural characteristics, strengths of all evaluated models are summarized below:

\noindent \textbf{Traditional region-based method:}
\begin{itemize}
    \item \textbf{ResNet-101}~\cite{kaiming2015resnet} with \textbf{Faster R-CNN}~\cite{ren2017FasterRcnn}. Faster R-CNN~\cite{ren2017FasterRcnn} is a classic two-stage detector that employs a ResNet-101~\cite{kaiming2015resnet} backbone with a Region Proposal Network to generate candidate object regions. It prioritizes accuracy and provides a strong baseline for layout detection.
    \item \textbf{ResNet-101}~\cite{kaiming2015resnet} with \textbf{Mask R-CNN}~\cite{He_2017_MaskRcnn}. Mask R-CNN~\cite{He_2017_MaskRcnn} is a two-stage, region-based detector that extends the Faster R-CNN~\cite{ren2017FasterRcnn} architecture by adding a parallel mask segmentation branch for each detected object.
    \item \textbf{ResNet-101}~\cite{kaiming2015resnet} with \textbf{DocSegTr}~\cite{biswas2022docsegtr}. DocSegTr~\cite{biswas2022docsegtr} is a self-attention integrated model for documents that combines visual features with a twin-attention module. As one of the first attention-driven approaches for document layout analysis, it emphasizes efficient semantic reasoning and achieves competitive accuracy on complex layouts.
    \item \textbf{DiT}~\cite{li2022dit} with \textbf{Cascade R-CNN}~\cite{cai2018cascade}. This combination integrates the Document Image Transformer (DiT)~\cite{li2022dit} as the backbone in a Cascade R-CNN~\cite{cai2018cascade} detection framework. The DiT~\cite{li2022dit} backbone, pre-trained on large-scale unlabeled document images, provides rich document-specific features, while the cascade of detectors, which is trained with increasing IoU thresholds, refines layout element predictions at multiple stages for improved quality.
    \item  \textbf{LayoutLMv3}~\cite{huang2022layoutlmv3} with \textbf{Cascade R-CNN}~\cite{cai2018cascade}. LayoutLMv3~\cite{huang2022layoutlmv3} is a pre-trained multimodal Transformer that jointly encodes text, layout coordinates, and image signals with unified masking strategies. We employ LayoutLMv3 as the backbone in a Cascade R-CNN – leveraging its cross-modal document representations – and refine predictions through multiple stages, which yields state-of-the-art accuracy in document layout analysis.
\end{itemize}

\noindent \textbf{DETR-based method:}
\begin{itemize}
    \item \textbf{InternImage}~\cite{wang2023internimage} with \textbf{DETR}~\cite{carion2020detr}. This combination pairs the large CNN-based InternImage~\cite{wang2023internimage} backbone with the DETR~\cite{carion2020detr} set-prediction framework for object detection. The backbone uses deformable convolutions to attain a large effective receptive field and adaptive spatial aggregation, complementing the end-to-end Transformer decoder of DETR~\cite{carion2020detr} to produce strong layout detection results without the need for non-maximum suppression.
    \item \textbf{InternImage}~\cite{wang2023internimage} with \textbf{Deformable DETR}~\cite{zhu2021deformableDETR}. This model combines with Deformable DETR~\cite{zhu2021deformableDETR}, an improved DETR variant that employs multi-scale deformable attention for faster convergence and better small-object detection. The image features together with efficient deformable attention allow the detector to learn document layouts more effectively, improving both training speed and detection accuracy over the  DETR~\cite{carion2020detr} approach.
    \item \textbf{InternImage}~\cite{wang2023internimage} with \textbf{DINO}~\cite{zhang2022dino}. In this combination, we use the DINO~\cite{zhang2022dino} framework as a detector, which is a DETR~\cite{carion2020detr} with improved denoising and query initialization techniques. By using contrastive denoising training and hybrid anchor box queries, DINO~\cite{zhang2022dino} enhances detection performance and, coupled with strong visual features, achieves high mAP on the document layout analysis task.
    \item \textbf{InternImage}~\cite{wang2023internimage} with \textbf{Co-DINO}~\cite{zong2022codetr}. This model builds upon Collaborative DETR~\cite{zong2022codetr} training schemes applied to the DINO~\cite{zhang2022dino} detector. It utilizes a collaborative hybrid label assignment strategy, \ie, training multiple auxiliary heads with one-to-many assignments to boost the learning of the model. The Co-DINO~\cite{zong2022codetr} approach strengthens the encoder’s feature extraction and improves query decoding, resulting in improved detection precision compared to the standard DINO~\cite{zhang2022dino}.
    \item \textbf{Swin Transformer}~\cite{liu2021swin} with \textbf{SwinDocSegmenter}~\cite{banerjee2023swindocsegmenter}. SwinDocSegmenter~\cite{banerjee2023swindocsegmenter} is an end-to-end transformer for document layout analysis that introduces a domain-adaptive training approach with contrastive learning and mixed query initialization in the transformer decoder, enabling robust analysis of complex document layouts and achieving superior accuracy.
    \item \textbf{InternImage}~\cite{wang2023internimage} with \textbf{RoDLA}~\cite{chen2024rodla}. RoDLA is a DINO-inspired~\cite{zhang2022dino} detector focused on robustness under image perturbations. It incorporates channel-attention and average-pooling blocks into the DINO~\cite{zhang2022dino} transformer encoder and uses the InternImage~\cite{wang2023internimage} backbone pre-trained on ImageNet-22K. This design enhances multi-scale feature stability, yielding a highly robust layout detector that achieves state-of-the-art performance on document layout analysis tasks even under challenging conditions.
    \item \textbf{ResNet-50}~\cite{kaiming2015resnet} with \textbf{DLAFormer}~\cite{wang2024dlaformerendtoendtransformerdocument}. DLAFormer~\cite{wang2024dlaformerendtoendtransformerdocument} is an end-to-end transformer model for Document Layout Analysis that unifies multiple sub-tasks in a single DETR-like~\cite{carion2020detr} framework. Using a ResNet-50~\cite{kaiming2015resnet} backbone for feature extraction, it treats detection, classification, and even reading order as joint relation predictions in a unified label space. This approach, along with type-specific query embeddings with text and layout modalities integration, allows DLAFormer to concurrently detect layout elements and infer their roles, achieving top performance. However, there is no detailed description of the model-specific internal module structure, and the code was not open-sourced after publication. We are unable to reproduce its structural design and can only compare it with the results provided in the paper for reference.
\end{itemize}

\noindent \textbf{Diffusion-based method:}
\begin{itemize}
    \item \textbf{ResNet-50}~\cite{kaiming2015resnet} with \textbf{DiffusionDet}~\cite{chen2022diffusiondet}. DiffusionDet~\cite{chen2022diffusiondet} formulates object detection as a denoising diffusion process that transforms noisy bounding boxes into accurate object boxes. Implemented with a ResNet-50~\cite{kaiming2015resnet} backbone, the model generates a set of random boxes and iteratively refines them through the diffusion model, allowing a flexible number of proposals and improved results with more refinement steps. This diffusion-based detector provides a novel paradigm for layout analysis by treating detection as progressive generation rather than direct regression.
    \item \textbf{Swin Transformer}~\cite{liu2021swin} with \textbf{DiffusionDet}~\cite{chen2022diffusiondet}. This combination applies the same diffusion-based detection approach using a Swin Transformer~\cite{liu2021swin} backbone for feature extraction. The stronger backbone captures richer visual context from document images, which improves the quality of the denoising process and yields higher detection accuracy. By combining the features with iterative diffusion refinement, this model achieves the best results among diffusion-based methods in our comparison.
\end{itemize}
\noindent \textbf{Autoregressive method:}
\begin{itemize}
    \item \textbf{ViT}~\cite{dosovitskiy2020vit} with \textbf{Pix2Seq}~\cite{chen2021pix2seq}. Pix2Seq~\cite{chen2021pix2seq} is an autoregressive object detection method that casts detection as a language modeling task, \ie, the model outputs a sequence of tokens representing bounding box coordinates and class labels. This combination uses a ViT~\cite{dosovitskiy2020vit} backbone to encode the input image into a sequence of tokens, then a Transformer decoder generates the sequence of layout element descriptions. This approach allows the model to learn to read out object locations and classes sequentially, yielding competitive detection results with a unified generative framework.
    \item \textbf{FIT}~\cite{chen2023fit} with \textbf{Pix2Seq}~\cite{chen2021pix2seq}. This combination employs the Far-reaching Interleaved Transformer (FIT)~\cite{chen2023fit} architecture as the backbone for the Pix2Seq~\cite{chen2021pix2seq} framework. FIT~\cite{chen2023fit} is a large Transformer that interleaves local and global self-attention layers, enabling efficient modeling of very long token sequences. By using a powerful encoder, this Pix2Seq~\cite{chen2021pix2seq} model can handle high-resolution document images and long output sequences more effectively.
\end{itemize}

\noindent \textbf{Our Proposed hybrid method}:
\begin{itemize}
    \item \textbf{ResNet‑50}~\cite{kaiming2015resnet} with \textbf{HybriDLA}. Using ResNet‑50~\cite{kaiming2015resnet} as a lightweight visual backbone, our HybriDLA detector retains the feature‑fusion encoder and hybrid generative decoder pipeline, allowing coarse‑to‑fine query expansion even on limited compute. 
    \item \textbf{ResNet‑101}~\cite{kaiming2015resnet} with \textbf{HybriDLA}. Replacing the backbone with a deeper ResNet‑101~\cite{kaiming2015resnet} provides richer convolutional features that are fed into the same hybrid decoder, improving spatial detail during diffusion‑style refinement. 
    \item  \textbf{ViT}~\cite{dosovitskiy2020vit} with \textbf{HybriDLA}. This variant utilizes a ViT~\cite{dosovitskiy2020vit} encoder, whose global self-attention is paired with query decoding from HybriDLA, allowing the model to capture long-range layout dependencies with minimal inductive bias.
    \item \textbf{Swin Transformer}~\cite{liu2021swin} with \textbf{HybriDLA}. Integrating a Swin Transformer~\cite{liu2021swin} backbone supplies multi‑scale window self‑attention features that align with our feature‑fusion encoder, yielding stronger localization for small layout elements.
    \item \textbf{InternImage}~\cite{wang2023internimage} with \textbf{HybriDLA}. Coupling HybriDLA with the large deformable‑convolution InternImage~\cite{wang2023internimage} backbone maximizes receptive‑field coverage and adaptive feature aggregation before hybrid decoding.
\end{itemize}
All models were trained under comparable settings for a fair evaluation with the Nvidia A100 GPUs. We trained all models for 36 epochs on both DocLayNet and M$^6$Doc. Training uses the AdamW optimizer with an initial learning rate of $2\times10^{-4}$ for a base batch size of 40 images and weight decay of $1\times10^{-4}$. The learning rate follows a multi-step schedule, dropping by $10\times$ at epoch 30. We employed gradient clipping with max norm 0.5 to stabilize training, and maintained an exponential moving average (EMA) of model weights with decay rate 0.9999 for more robust convergence. 

\section{Additional Detailed Analysis of Ablation Study}
To further validate our design choices, we conduct two additional sets of ablation experiments, focusing on the influence of architectural components and hyperparameters on detection performance. In particular, we examine the effect of the Transformer decoder depth together with our diffusion-based refinement (DR) mechanism and feature fusion encoder (FFE), and we investigate the interplay between the encoder type, the presence of the DR module, and the number of queries used in the autoregressive query expansion (AQE) pipeline. All experiments are performed on the DocLayNet~\cite{pfitzmann2022doclaynet} dataset using two backbone architectures, \ie, ResNet-50~\cite{kaiming2015resnet} and Swin Transformer~\cite{liu2021swin}, to observe these effects.

\begin{table}[h]
\centering
\caption{Impact of number of decoder layers under different combinations of the diffusion-based refinement (DR) mechanism with the feature fusion encoder (FFE). All models use 300 initial queries and 300 AQE queries on the DocLayNet~\cite{pfitzmann2022doclaynet} dataset.}
\label{tab:decoder_depth_ablation}
\begin{adjustbox}{width=\linewidth}
\begin{tabular}{ccccccc}
\toprule
Backbone & Detector & \makecell{Decoder\\Layers} & Encoder & DR & AQE & mAP\\
\midrule
ResNet-50 & HybriDLA & 1 & FFE & \xmark & \cmark & 74.20 \\
ResNet-50 & HybriDLA & 1 & FFE & \cmark & \cmark & 72.80 \\
ResNet-50 & HybriDLA & 3 & FFE & \xmark & \cmark & 74.40 \\
ResNet-50 & HybriDLA & 3 & FFE & \cmark & \cmark & 73.10 \\
ResNet-50 & HybriDLA & 6 & FFE & \xmark & \cmark & 74.40 \\
ResNet-50 & HybriDLA & 6 & FFE & \cmark & \cmark & 74.40 \\
\midrule
Swin-L    & HybriDLA & 1 & FFE & \xmark & \cmark & 76.80 \\
Swin-L    & HybriDLA & 1 & FFE & \cmark & \cmark & 76.60 \\
Swin-L    & HybriDLA & 3 & FFE & \xmark & \cmark & 78.90 \\
Swin-L    & HybriDLA & 3 & FFE & \cmark & \cmark & 78.20 \\
Swin-L    & HybriDLA & 6 & FFE & \xmark & \cmark & 79.10 \\
Swin-L    & HybriDLA & 6 & FFE & \cmark & \cmark & 80.40 \\
\bottomrule
\end{tabular}
\end{adjustbox}
\end{table}
\noindent \textbf{The ablation study of the decoder depth.}
Table~\ref{tab:decoder_depth_ablation} examines how the decoder depths influence mAP under different settings of our diffusion-based refinement (DR) mechanism with the feature fusion encoder fixed. We report results on DocLayNet for a lightweight ResNet-50~\cite{kaiming2015resnet} backbone with a large Swin Transformer~\cite{liu2021swin} backbone, each evaluated with 1, 3, and 6 decoder layers.

For ResNet-50~\cite{kaiming2015resnet}, increasing the decoder depth from 1 to 3 layers yields a marginal gain, \ie, mAP from 74.2\% to 74.4\% without DR, and adding more layers beyond that shows no further improvement from 3 to 6 layers, remaining at 74.4\% mAP. Interestingly, enabling the DR mechanism with an insufficient decoder depth can hurt performance. For example, with only 1 decoder layer, the model with DR actually performs worse than the one without DR. At 3 layers, the DR variant still lags behind. Only when the decoder is deep enough, \ie, 6 layers, does the ResNet-50 model fully recover the performance with DR. This suggests that for a smaller backbone, the diffusion refinement introduces extra complexity that a shallow decoder cannot effectively capitalize on. For Swin Transformer~\cite{liu2021swin}, the backbone consistently benefits from deeper decoders, and it can more effectively leverage the DR mechanism when enough decoder layers are present. Without DR, increasing decoder layers from 1 to 6 yields steady improvements. With DR, a single decoder layer is not yet advantageous, and 3 layers with DR reach 78.2\% mAP. Crucially, at 6 decoder layers, the Swin Transformer~\cite{liu2021swin} with DR beats the model without DR, achieving 80.4\% mAP. This indicates that when paired with a powerful backbone, the diffusion-based refinement mechanism can significantly improve performance. Overall, the ablation highlights that more transformer decoder layers generally improve detection performance, and the effectiveness of the DR mechanism is contingent on model capacity. It can slightly degrade results if the decoder is too shallow, especially on a weaker backbone, but becomes highly beneficial with a stronger backbone and a deeper decoder.

\begin{table}[h]
\centering
\caption{Performance under different architectural components and query numbers on DocLayNet~\cite{pfitzmann2022doclaynet}. 
Each row specifies the encoder type (feature fusion encoder (FFE) or deformable encoder (DE)), diffusion-based refinement (DR) mechanism, and autoregressive query expansion (AQE) mechanism, along with the number of initial and AQE queries.}
\label{tab:module-query-interaction}
\begin{adjustbox}{width=\linewidth}
\begin{tabular}{l c c c c c c}
\toprule
\textbf{Backbone} & \textbf{Encoder} & \textbf{DR} & \textbf{AQE} & \textbf{\#Init.\ Q.} & \textbf{\#AQE Q.} & \textbf{mAP} \\
\midrule
ResNet-50 & DE  & \cmark & \cmark & 300 & 300 & 76.2 \\
ResNet-50 & FFE & \xmark & \cmark & 300 & 300 & 74.4 \\
ResNet-50 & FFE & \cmark & \cmark & 300 & 300 & 74.4 \\
ResNet-50 & DE  & \cmark & \cmark & 300 & 30 & 77.8 \\
ResNet-50 & FFE & \xmark & \cmark & 300 & 30 & 76.3 \\
ResNet-50 & FFE & \cmark & \cmark & 300 & 30 & 77.4 \\
ResNet-50 & DE  & \cmark & \cmark & 900 & 30 & 74.2 \\
ResNet-50 & FFE & \xmark & \cmark & 900 & 30 & 73.5 \\
ResNet-50 & FFE & \cmark & \cmark & 900 & 30 & 75.5 \\

\midrule
Swin-L    & DE  & \cmark & \cmark & 300 & 300 & 78.1 \\
Swin-L    & FFE & \xmark & \cmark & 300 & 300 & 79.1 \\
Swin-L    & FFE & \cmark & \cmark & 300 & 300 & 80.4 \\
Swin-L    & DE  & \cmark & \cmark & 300 & 30 & 75.3 \\
Swin-L    & FFE & \xmark & \cmark & 300 & 30 & 77.6 \\
Swin-L    & FFE & \cmark & \cmark & 300 & 30 & 78.2 \\
Swin-L    & DE  & \cmark & \cmark & 900 & 30 & 76.7 \\
Swin-L    & FFE & \xmark & \cmark & 900 & 30 & 75.7 \\
Swin-L    & FFE & \cmark & \cmark & 900 & 30 & 77.3 \\

\bottomrule
\end{tabular}
\end{adjustbox}
\end{table}
\noindent \textbf{The ablation study of the interplay between the architectural components and hyperparameters.}
As shown in Table~\ref{tab:module-query-interaction}, we evaluated detection mAP on DocLayNet under various encoder configurations and query counts, comparing the deformable encoder (DE) baseline to our feature fusion encoder (FFE) with and without the diffusion-based refinement (DR) and autoregressive query expansion (AQE).
On the ResNet‑50~\cite{kaiming2015resnet} backbone, the deformable encoder that integrates DR yields the highest accuracy unless the FFE is also coupled with DR. When DR is enabled, FFE attains parity and in several settings overtakes the deformable encoder, demonstrating that iterative refinement can compensate for weaker geometric modeling on a light backbone. On the Swin Transformer~\cite{liu2021swin} backbone, the situation is reversed. FFE alone already beats the deformable encoder, and when DR is counted, the FFE becomes the clear optimum. The richer multi‑scale representations of Swin Transformer~\cite{liu2021swin} allow FFE to exploit DR more effectively than the deformable counterpart. Besides, a large pool of initial queries, \eg, 900, consistently lowers mean average precision for both backbones. Insufficient capacity in the decoder to process this overload causes optimization difficulties that DR can only partially help with. Empirically, assigning approximately 300 initial queries and 300 autoregressively expanded queries gives the most stable and accurate results, balancing coverage with computational tractability. Last, the benefit of DR is strongly correlated with overall model size. Deeper decoders and stronger backbones extract more value from the iterative refinement of queries, while shallow models show limited and negative returns when DR is introduced preemptively. 

\section{Quantitative Result Analysis and Failure Case}
\begin{figure*}[h]
\centering
% ---------- 
\begin{subfigure}[t]{0.49\textwidth}
  \includegraphics[width=\linewidth]{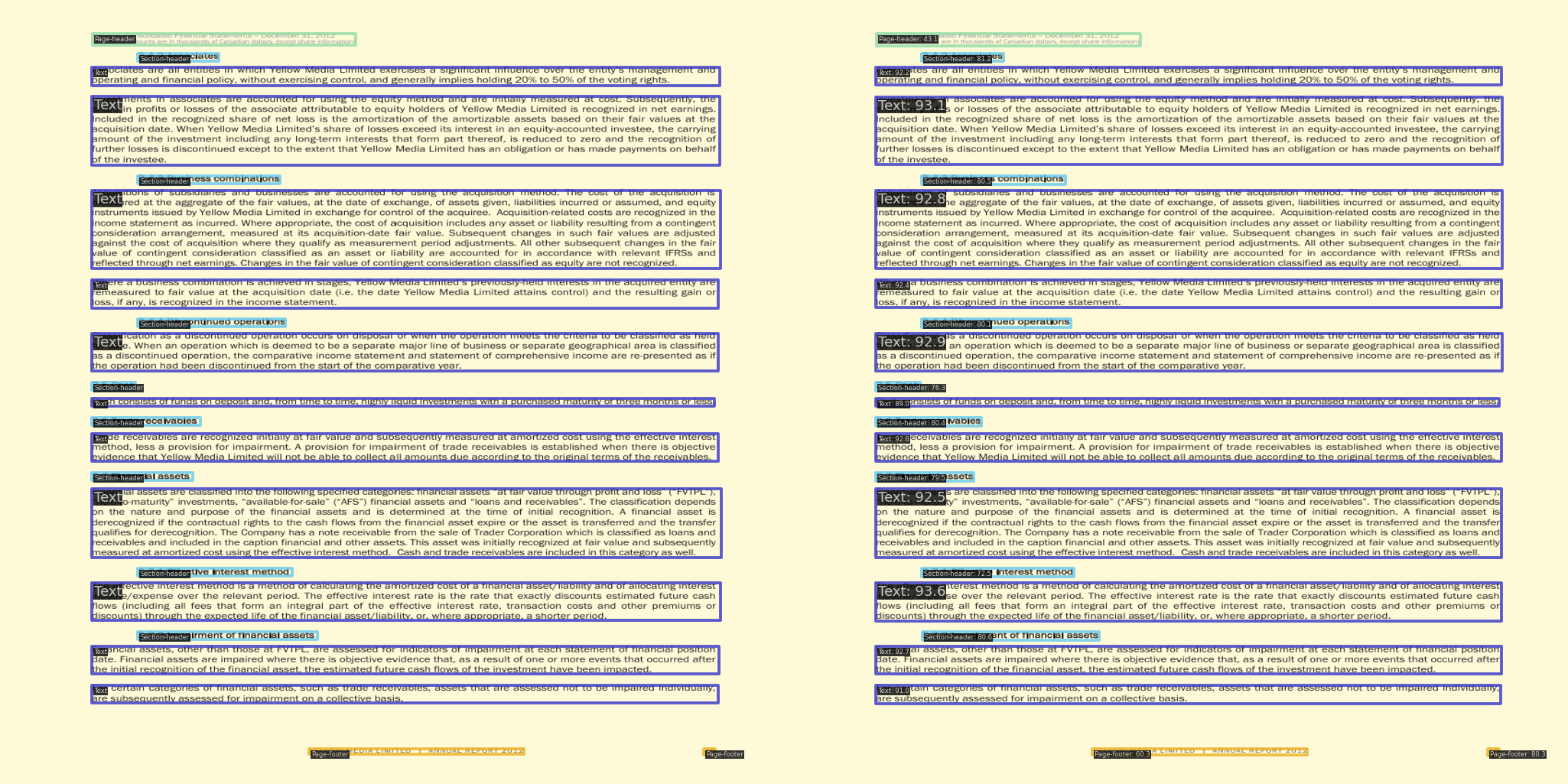}
  \label{fig:ex1}
\end{subfigure}\hfill
\begin{subfigure}[t]{0.49\textwidth}
  \includegraphics[width=\linewidth]{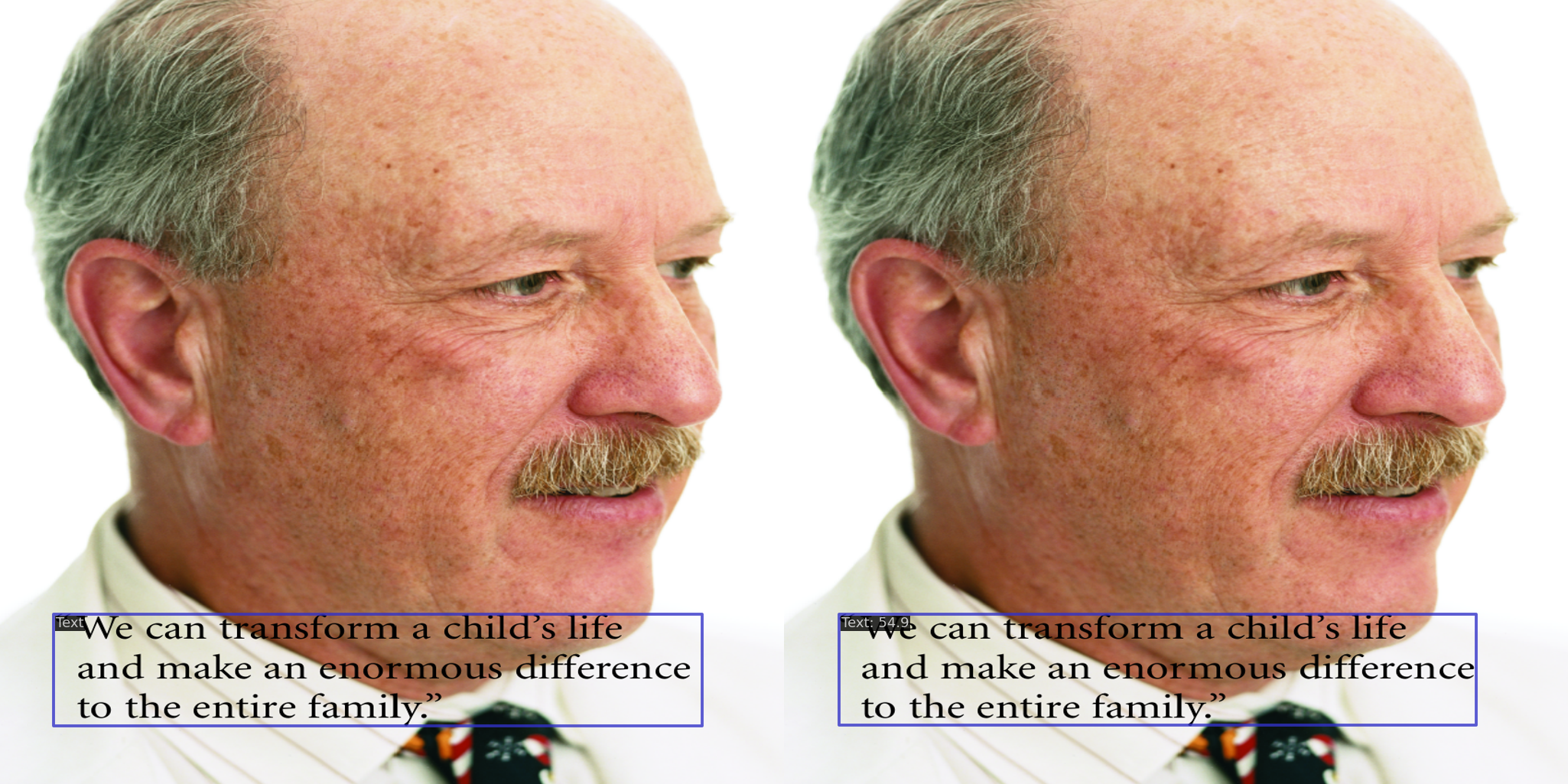}
  \label{fig:ex2}
\end{subfigure}

\vspace{4pt} 
% ---------- 
\begin{subfigure}[t]{0.49\textwidth}
  \includegraphics[width=\linewidth]{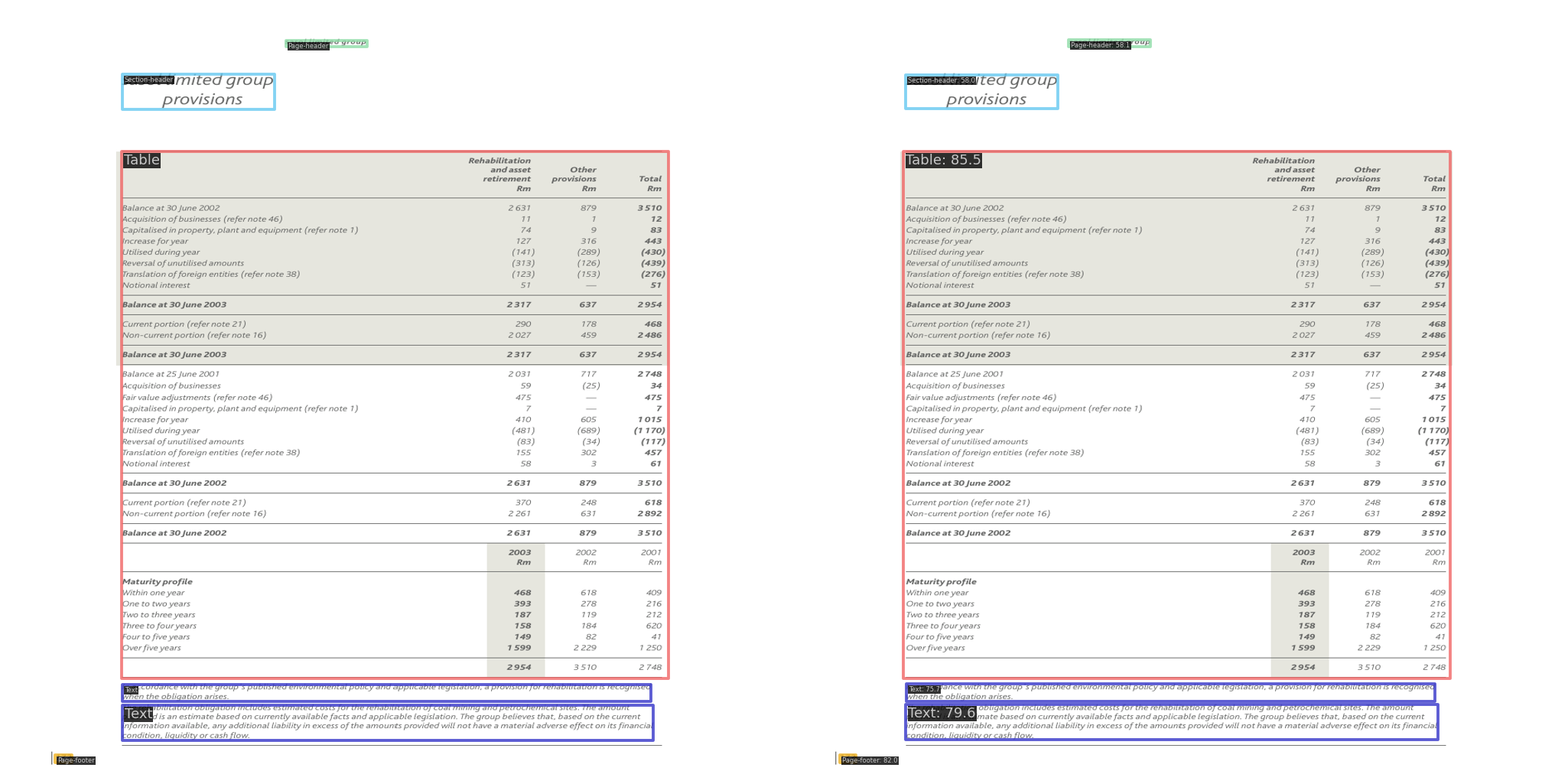}
  \label{fig:ex3}
\end{subfigure}\hfill
\begin{subfigure}[t]{0.49\textwidth}
  \includegraphics[width=\linewidth]{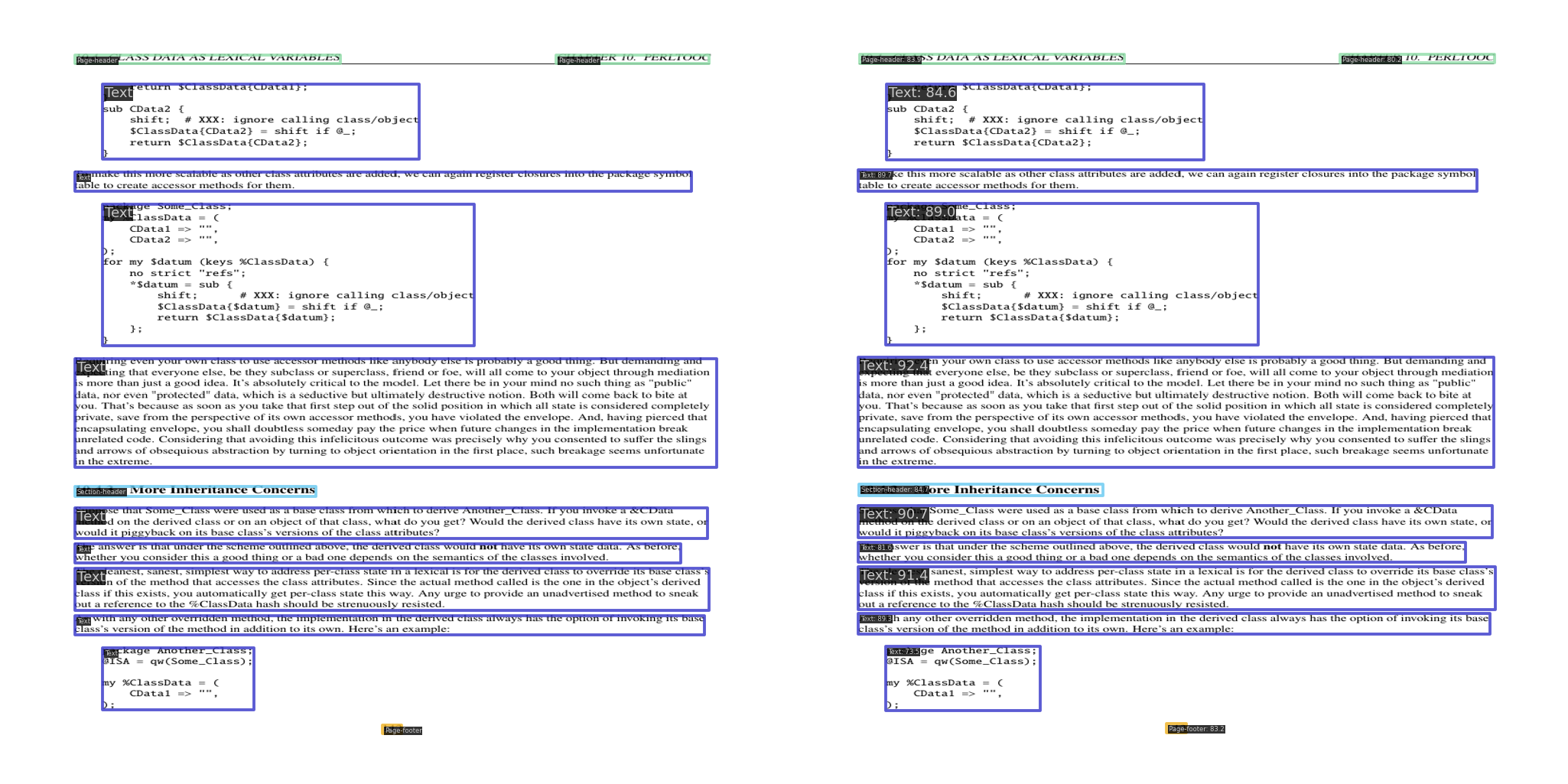}
  \label{fig:ex4}
\end{subfigure}

\vspace{4pt}
% ---------- 
\begin{subfigure}[t]{0.49\textwidth}
  \includegraphics[width=\linewidth]{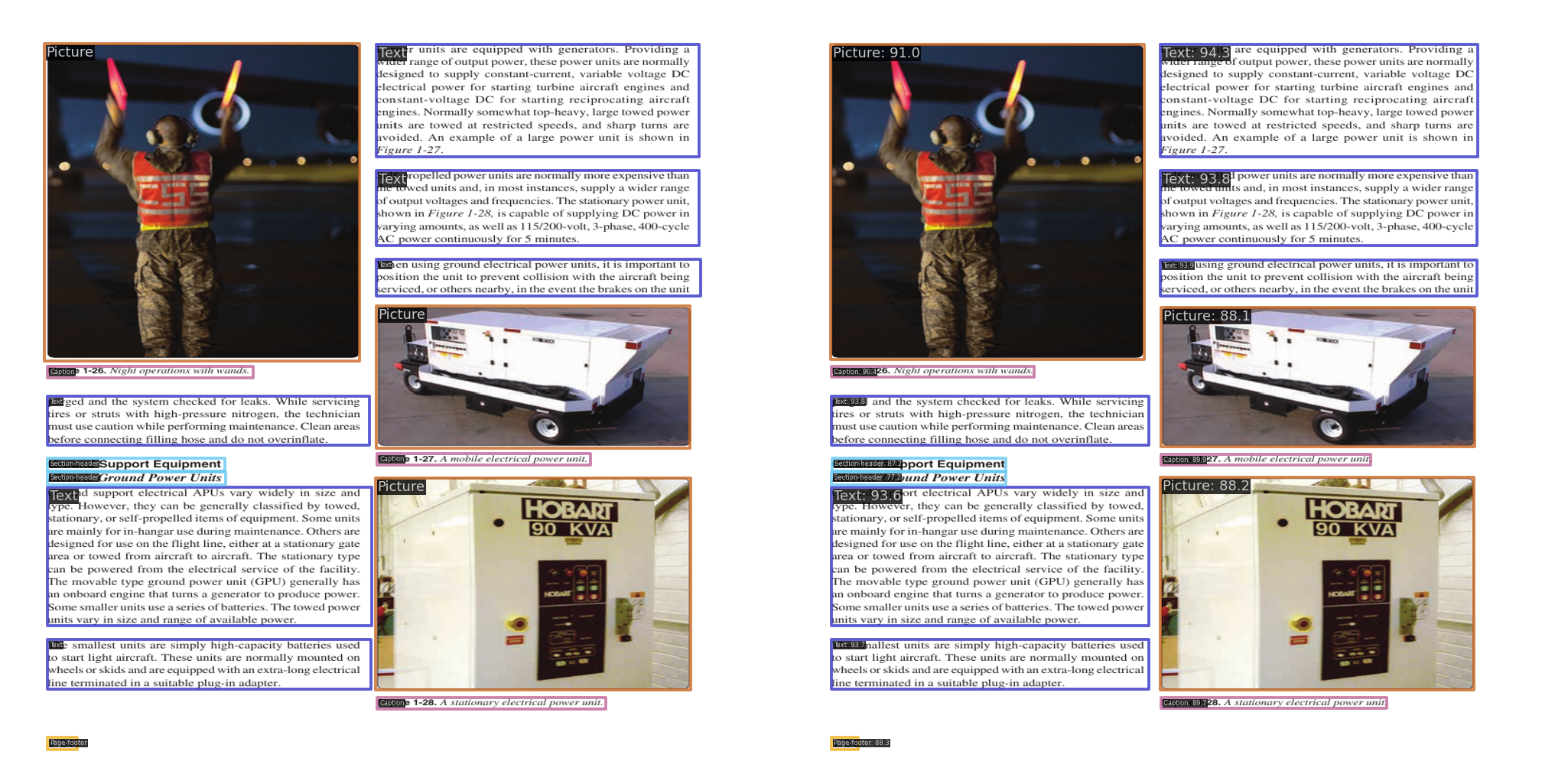}
  \label{fig:ex5}
\end{subfigure}\hfill
\begin{subfigure}[t]{0.49\textwidth}
  \includegraphics[width=\linewidth]{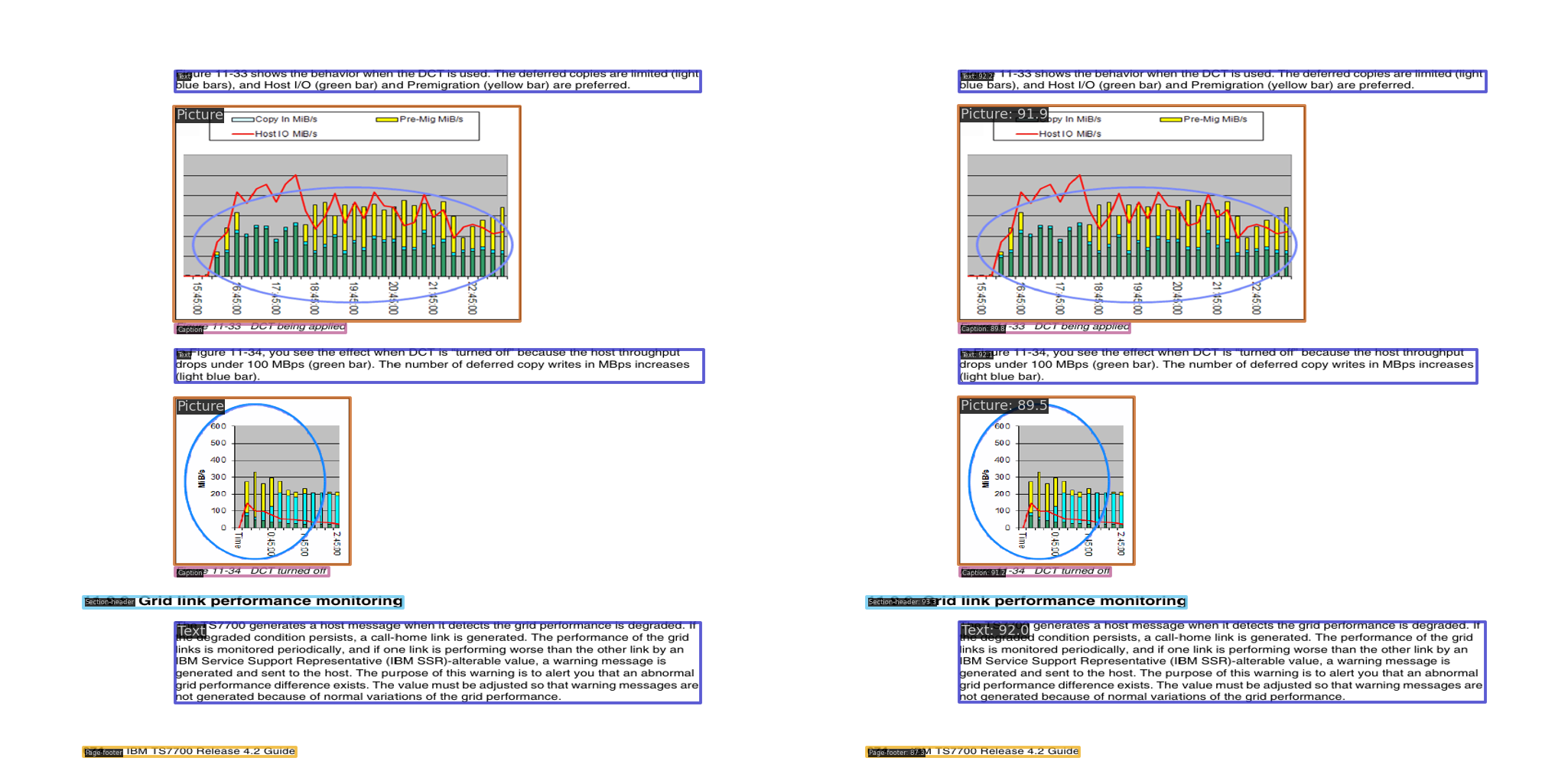}
  \label{fig:ex6}
\end{subfigure}

\vspace{4pt}
% ---------- 
\begin{subfigure}[t]{0.49\textwidth}
  \includegraphics[width=\linewidth]{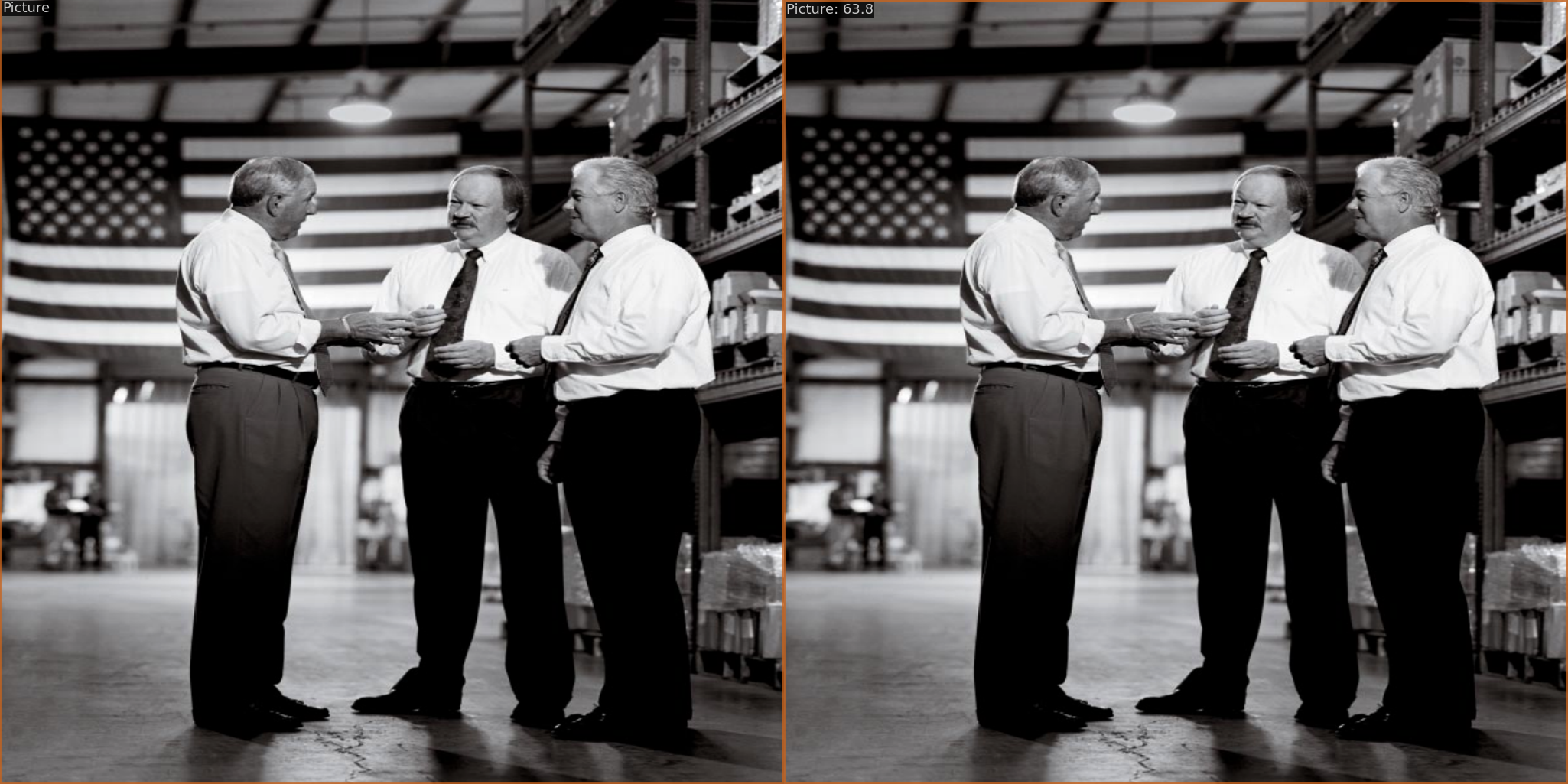}
  \label{fig:ex8}
\end{subfigure}\hfill
\begin{subfigure}[t]{0.49\textwidth}
  \includegraphics[width=\linewidth]{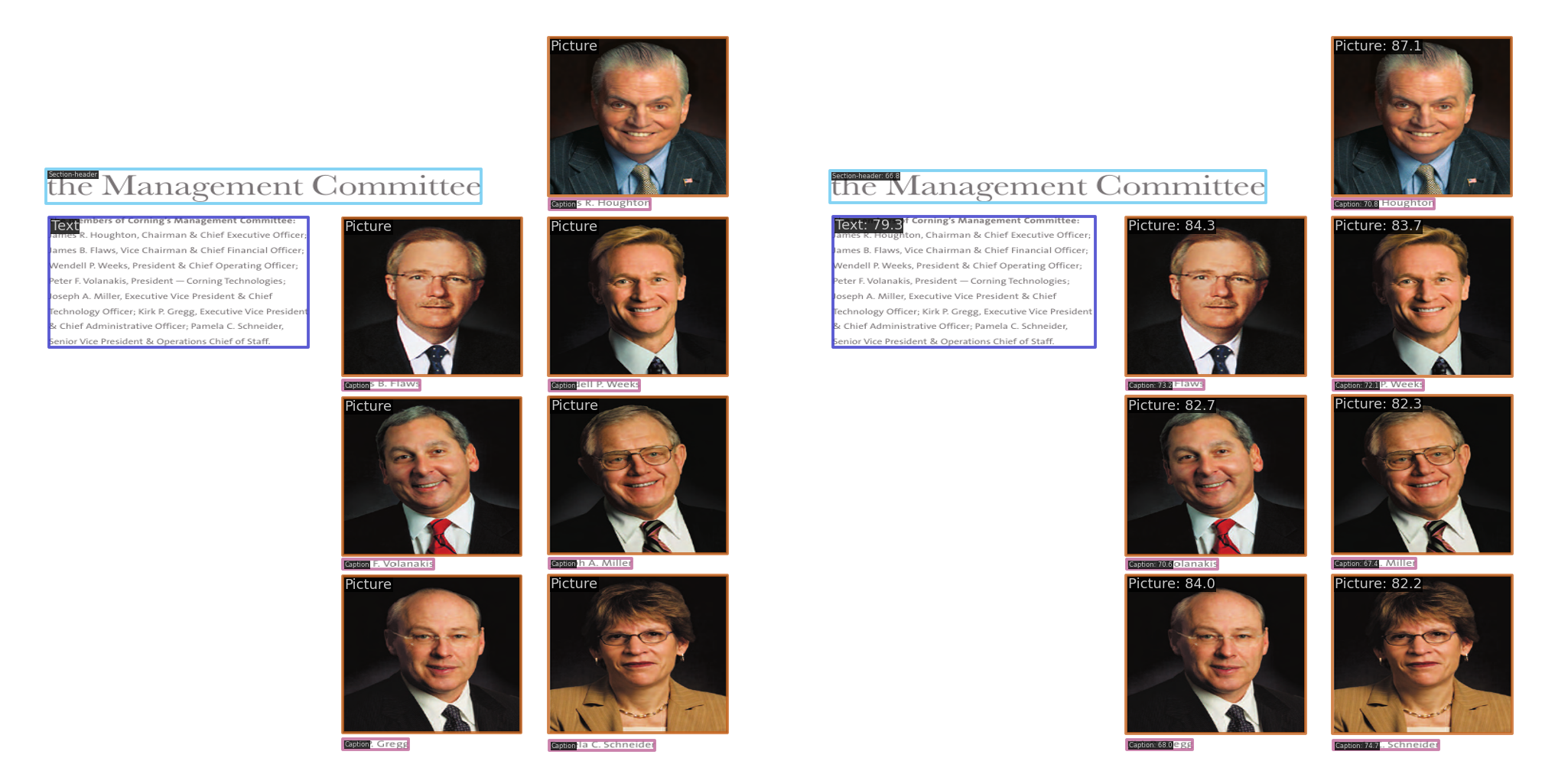}
  \label{fig:ex10}
\end{subfigure}

\caption{\textbf{HybriDLA with InternImage~\cite{wang2023internimage} Results on DocLayNet~\cite{pfitzmann2022doclaynet}.} Each subfigure shows a document page with ground-truth annotations on the left and the results of the model on the right. }
\label{fig:hybridla-doclaynet}
\end{figure*}
\subsection{Quantitative Result Analysis.}
As shown in Figure~\ref{fig:hybridla-doclaynet}, the HybriDLA method accurately identifies diverse document layout elements such as Page-header, Caption, Picture, Section-header, Table, and Text in complex layouts. It demonstrates robust performance across different document styles, including scientific articles, scanned newspaper pages, and black-and-white documents, highlighting its high detection precision and adaptability to varied layouts.
\subsection{Failure Case.}
\begin{figure*}[t]
\centering
% ---------- Row 1 ----------
\begin{subfigure}[t]{0.49\textwidth}
  \includegraphics[width=\linewidth]{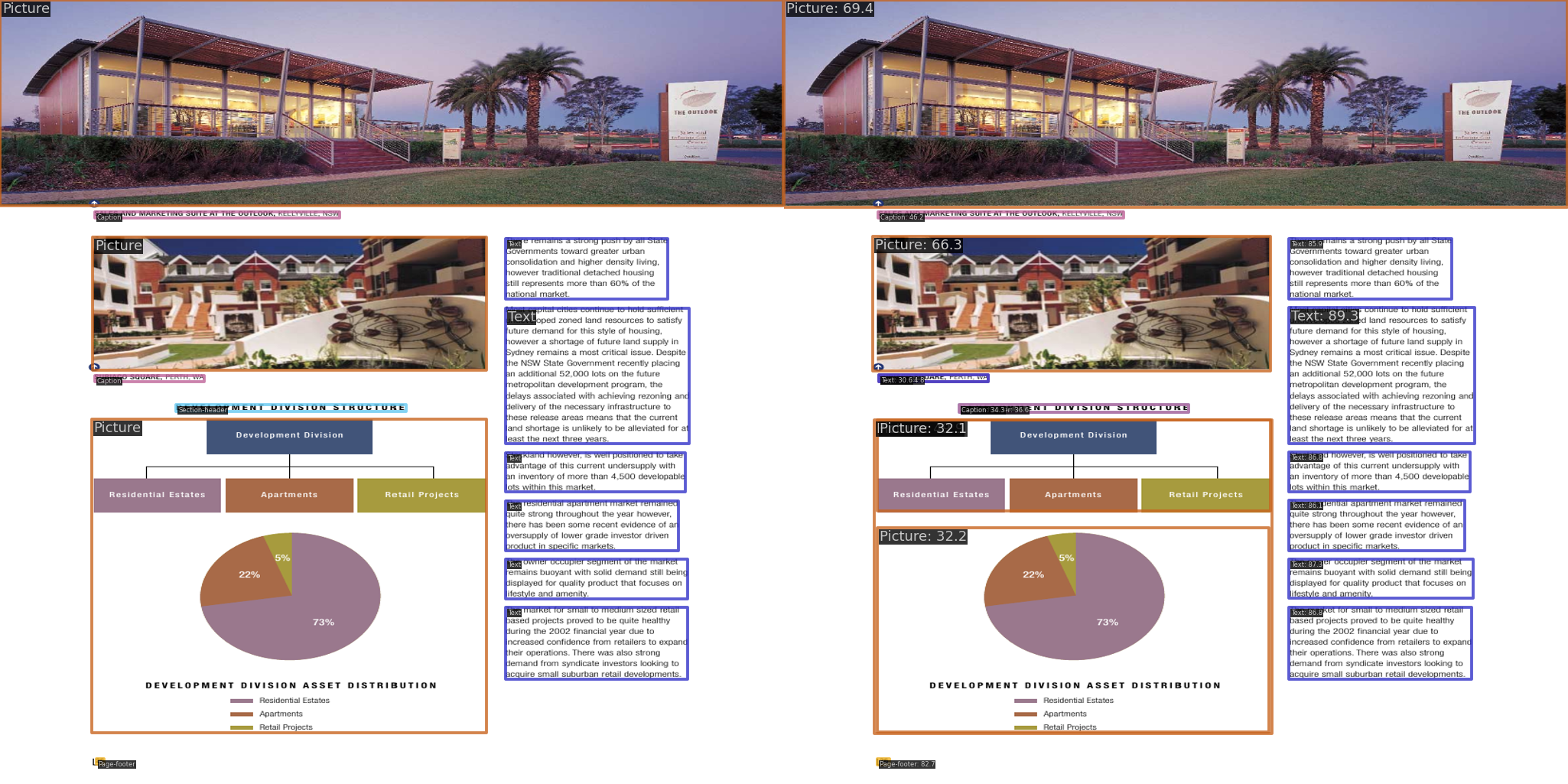}
  \label{fig:fail1}
\end{subfigure}\hfill
\begin{subfigure}[t]{0.49\textwidth}
  \includegraphics[width=\linewidth]{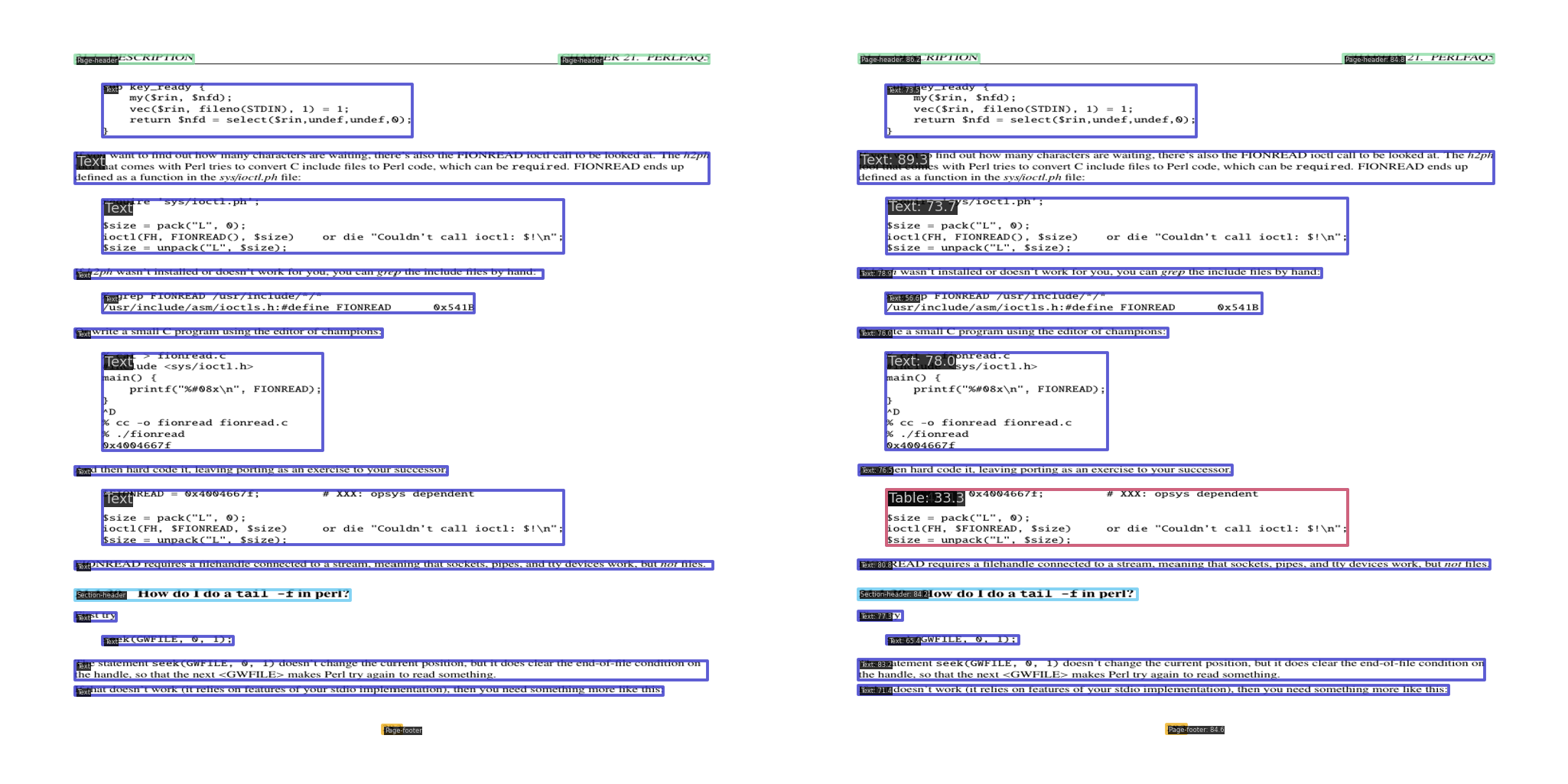}
  \label{fig:fail2}
\end{subfigure}

\vspace{4pt}
% ---------- Row 2 ----------
\begin{subfigure}[t]{0.49\textwidth}
  \includegraphics[width=\linewidth]{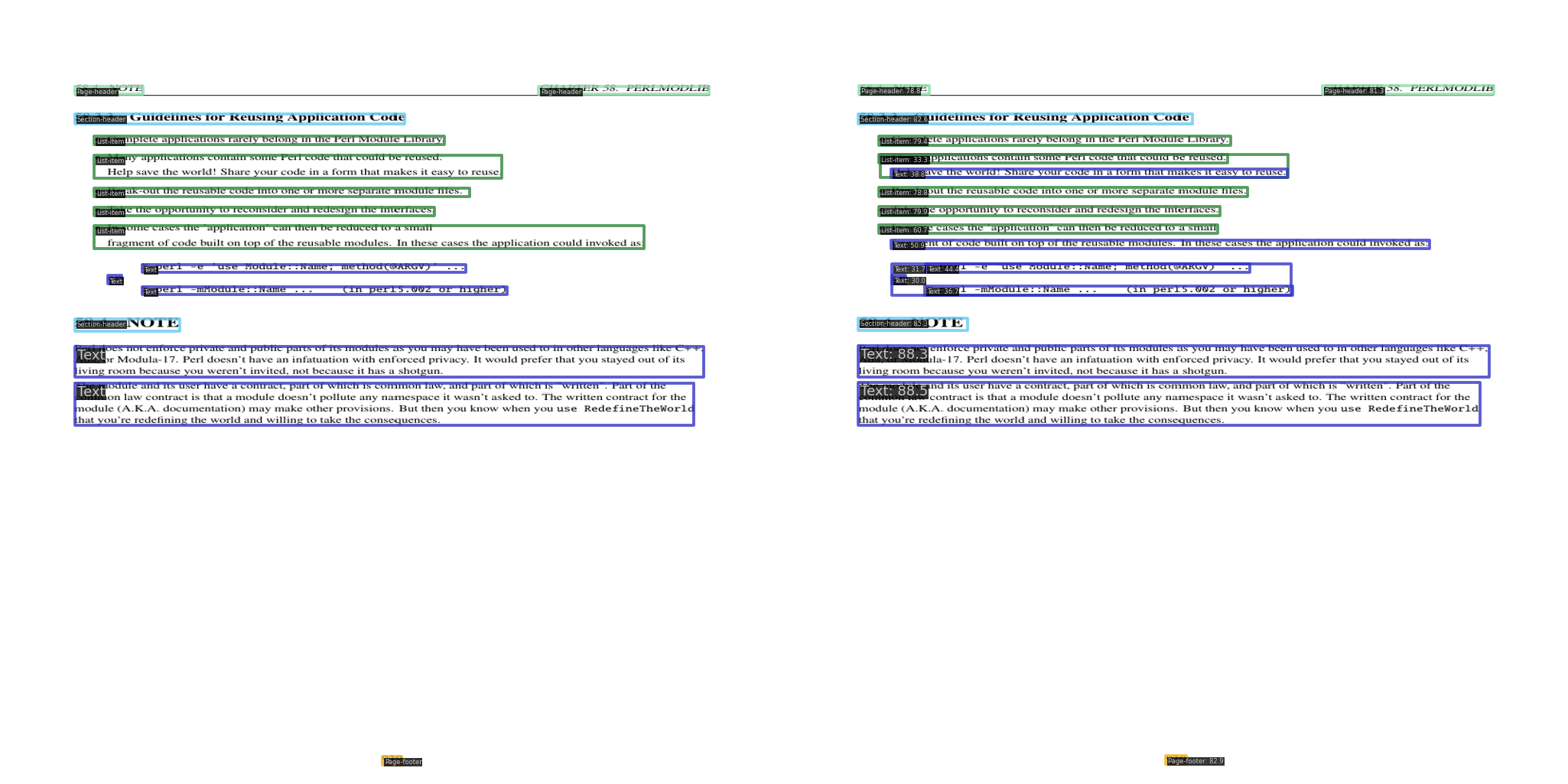}
  \label{fig:fail3}
\end{subfigure}\hfill
\begin{subfigure}[t]{0.49\textwidth}
  \includegraphics[width=\linewidth]{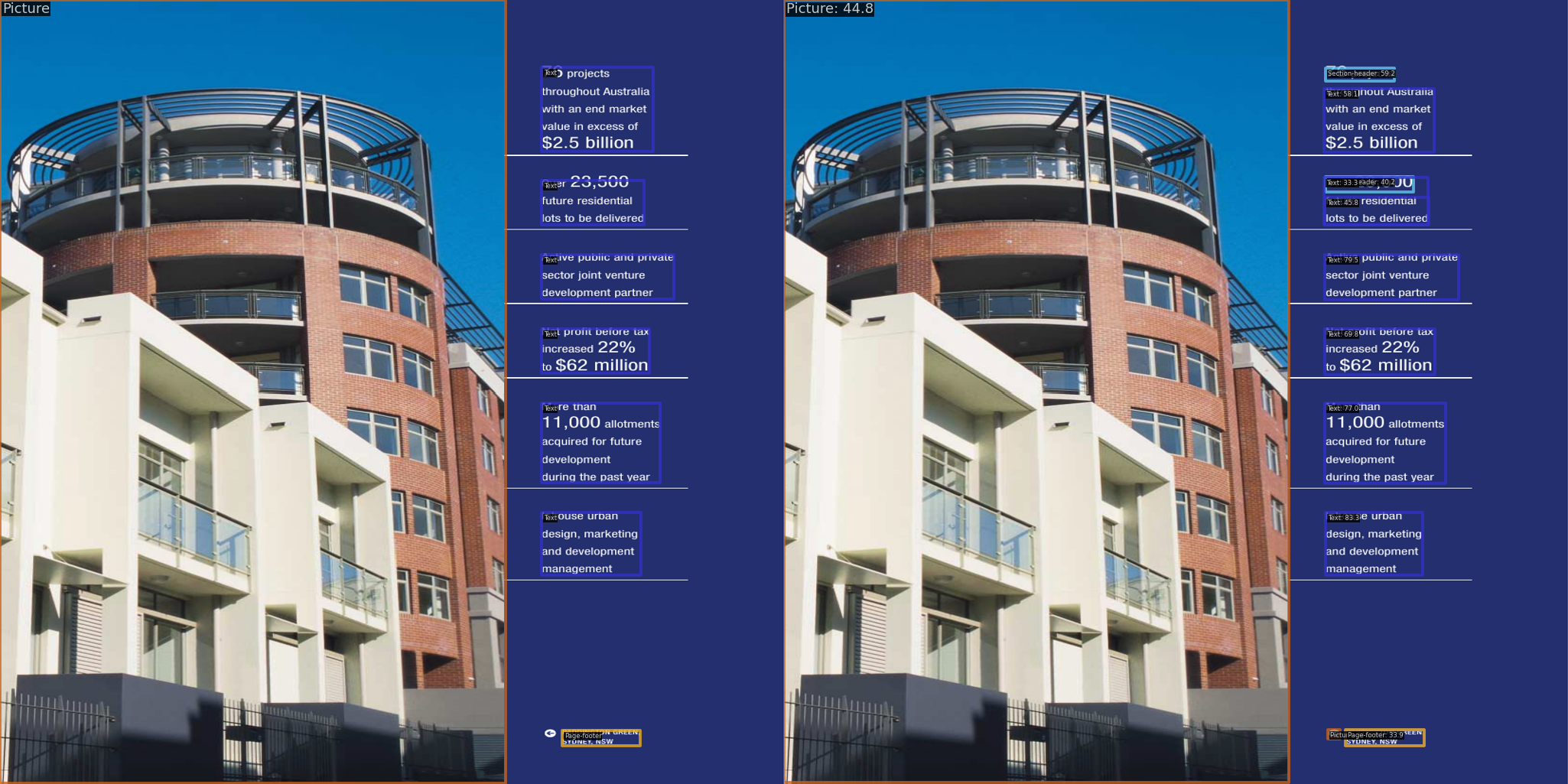}
  \label{fig:fail4}
\end{subfigure}

\vspace{4pt}
% ---------- Row 3 ----------
\begin{subfigure}[t]{0.49\textwidth}
  \includegraphics[width=\linewidth]{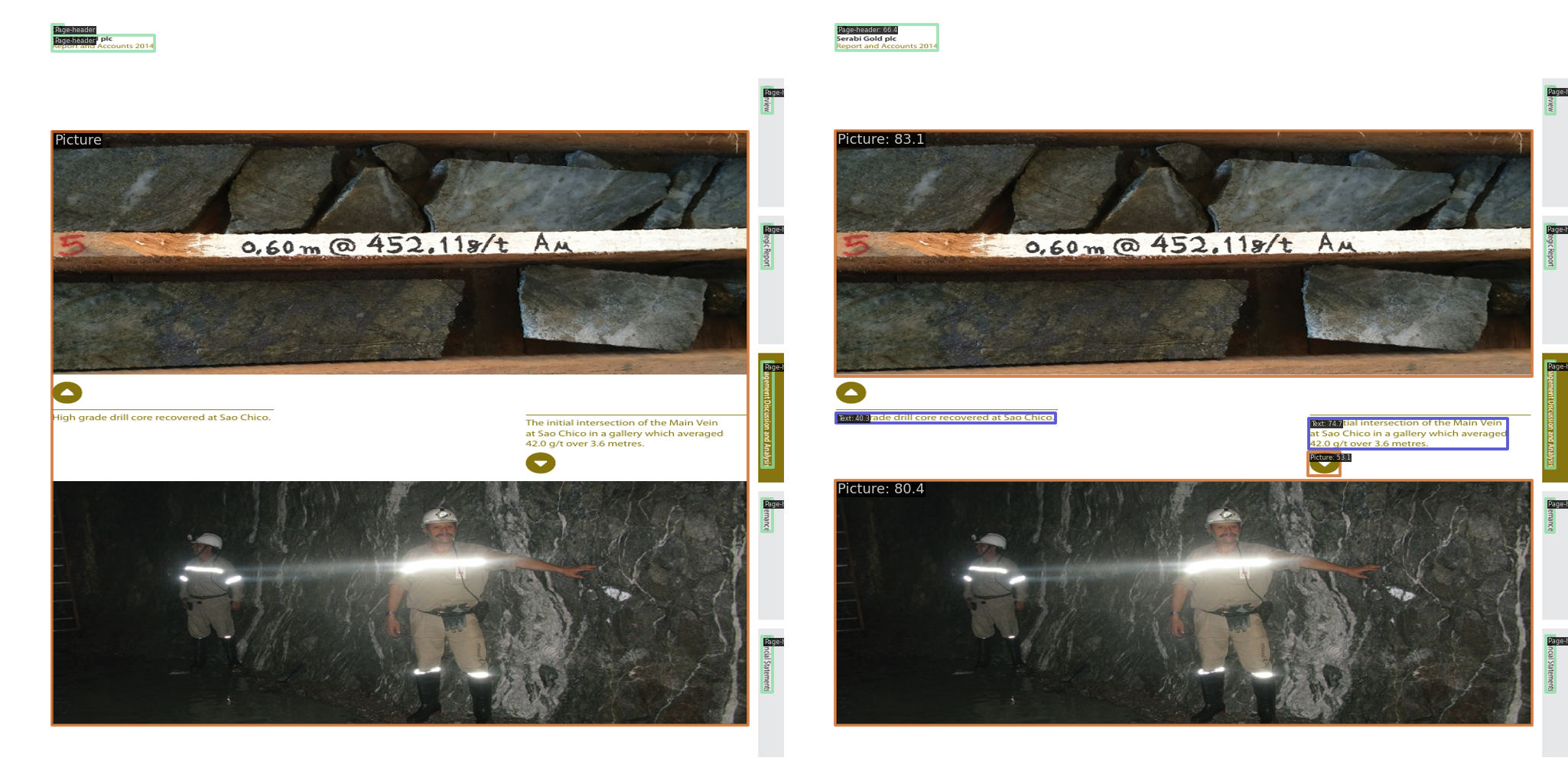}
  \label{fig:fail5}
\end{subfigure}\hfill
\begin{subfigure}[t]{0.49\textwidth}
  \includegraphics[width=\linewidth]{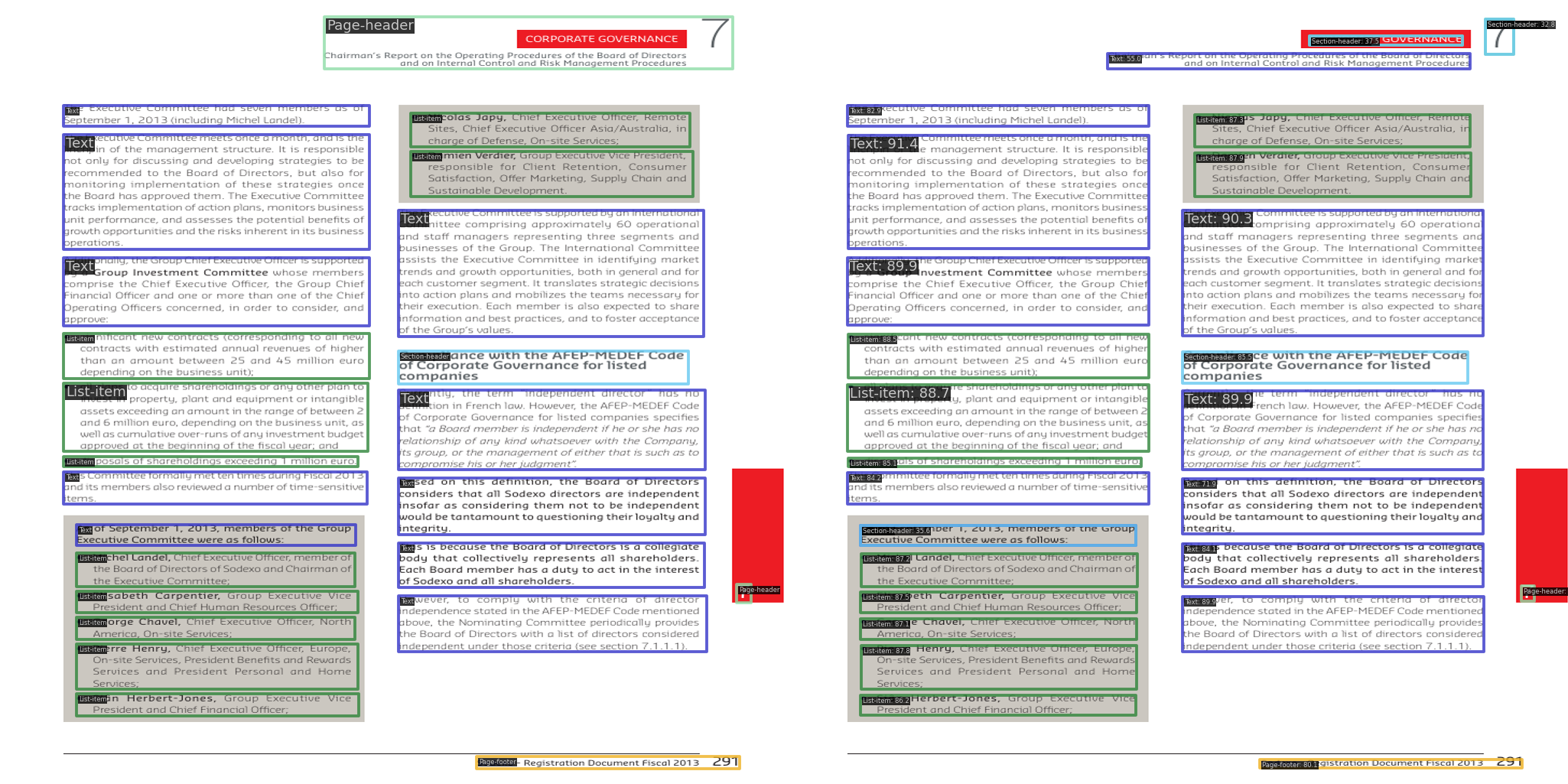}
  \label{fig:fail6}
\end{subfigure}

\vspace{4pt}
% ---------- Row 4 ----------
\begin{subfigure}[t]{0.49\textwidth}
  \includegraphics[width=\linewidth]{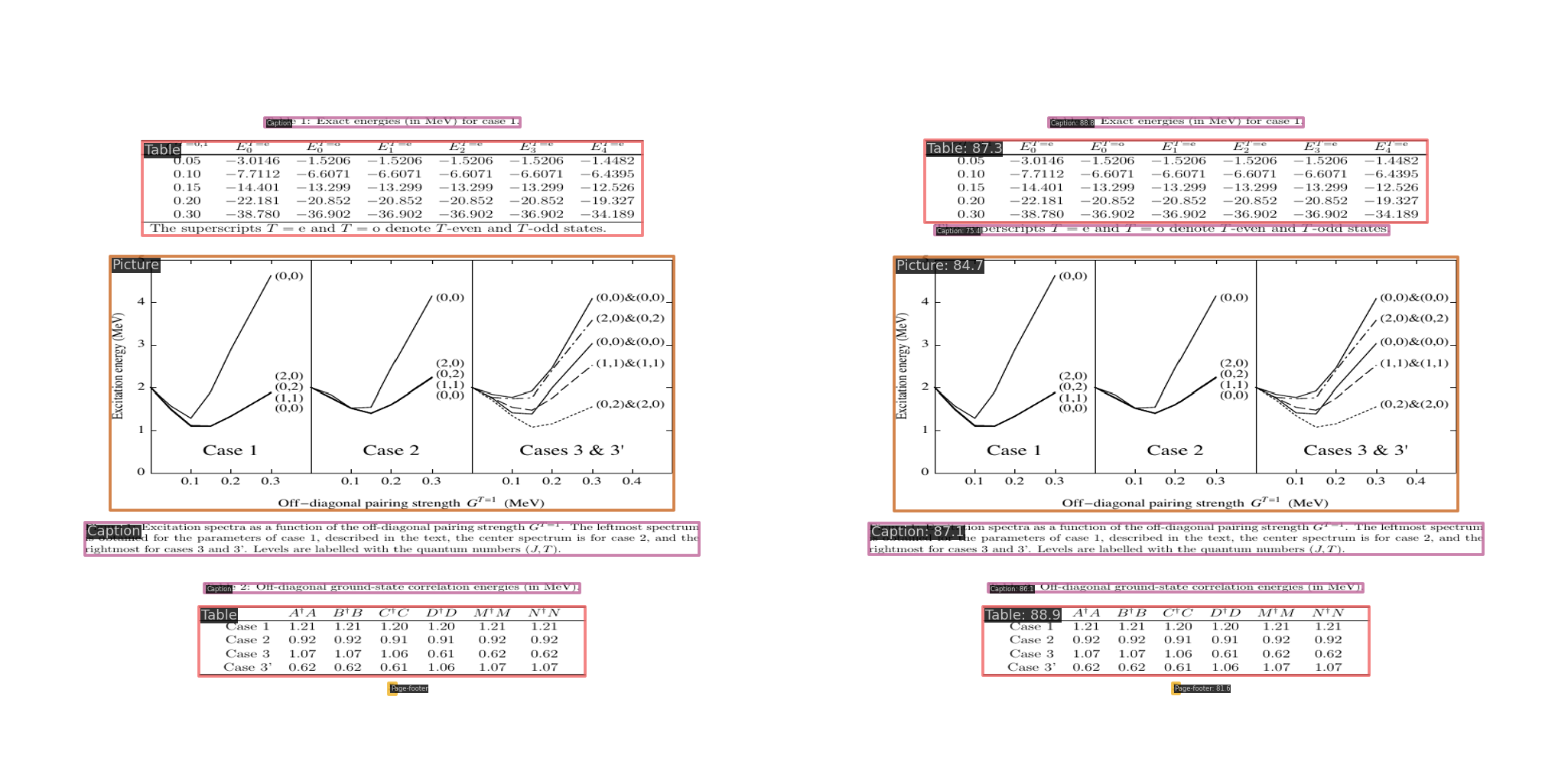}
  \label{fig:fail7}
\end{subfigure}\hfill
\begin{subfigure}[t]{0.49\textwidth}
  \includegraphics[width=\linewidth]{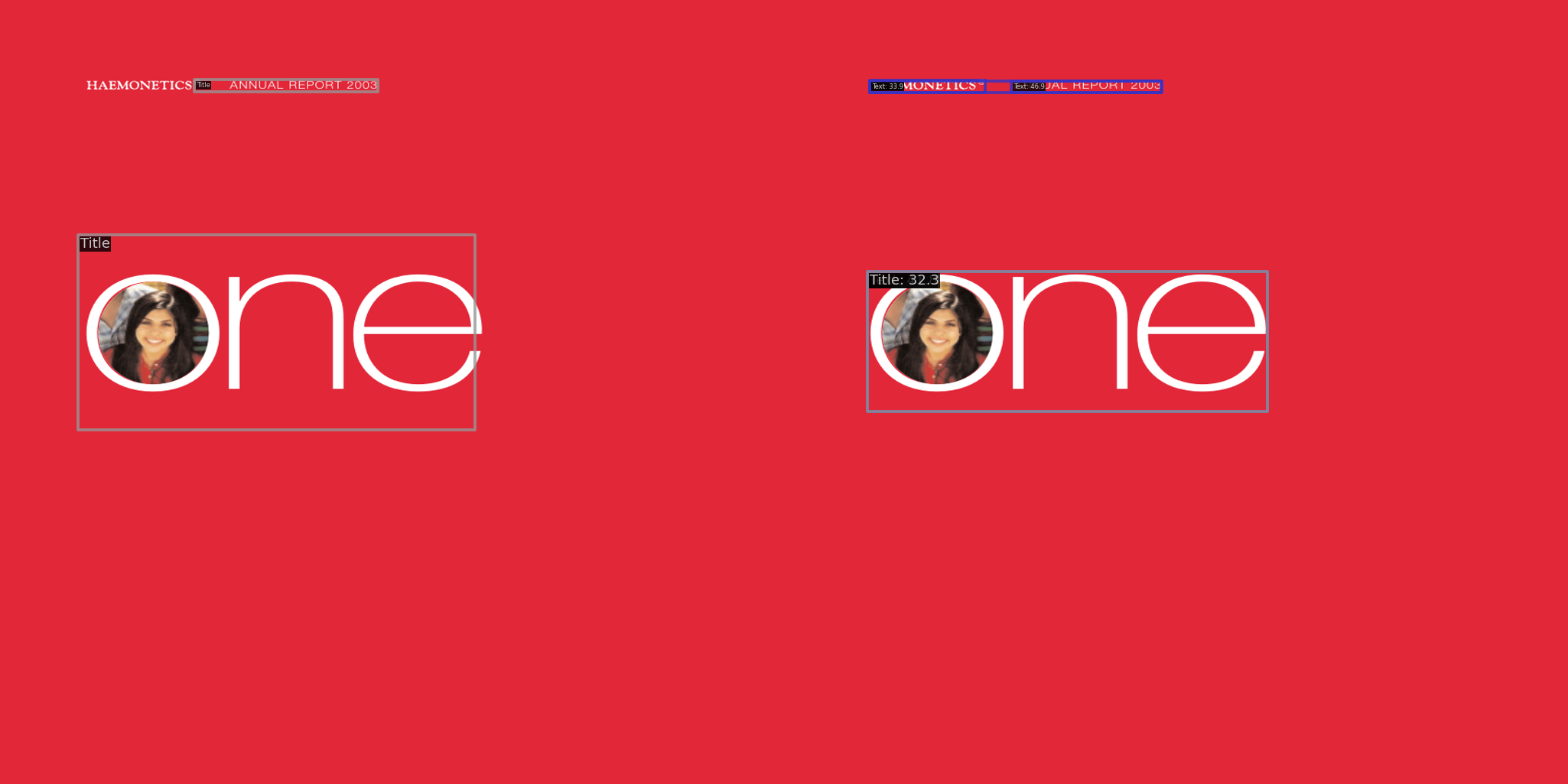}
  \label{fig:fail8}
\end{subfigure}

\caption{\textbf{Typical failure cases of HybriDLA with InternImage~\cite{wang2023internimage} on DocLayNet~\cite{pfitzmann2022doclaynet}}
For each page, the left half shows ground‑truth annotations, and the right half shows prediction results.}
\label{fig:hybridla-failure}
\end{figure*}
Figure~\ref{fig:hybridla-failure} gathers 8 representative failure pages and reveals three frequent patterns:

\begin{enumerate}[label=\roman*)]
    \item \textbf{Correct detections that are not present in the ground‑truth.}  
          In several pages, our detector identifies legitimate layout elements, \eg, small Page‑header banners or thin rules that separate a figure and its caption that were omitted by the human annotators in DocLayNet. These show up as false positives in quantitative scores, but are correct predictions, highlighting an upper‑bound limitation imposed by imperfect annotations.
    \item \textbf{Over‑refinement with too fragmented boxes.}  
          The hybrid diffusion stage occasionally produces multiple highly overlapping or over-divided boxes for one logical element, especially around dense text blocks and when a picture contains textual graphics. This is mainly from the iterative diffusion-based refinement steps.
    \item \textbf{Dominant background and atypical layouts.}  
          \begin{itemize}[leftmargin=*]
              \item \textbf{Full‑bleed photographs with embedded text}. Sparse textual regions on top of large images confuse the class priors and lead to caption and section‑header swaps or misses.  
              \item \textbf{Code snippets and dense marginalia}. Extremely small mono-spaced text blocks are mistaken for tables or captions.
              \item \textbf{Overlapping translucent layers or cross‑page elements}. Transparent overlays invalidate positional cues, yielding duplicate detections and missed list-items.
          \end{itemize}
\end{enumerate}

These cases indicate that a portion of the apparent errors are due to incomplete annotations, the diffusion‑based refinement can over‑divide in visually cluttered areas, and extremely unbalanced foreground–background ratios or non‑Manhattan layouts remain challenging. 

\end{document}